\documentclass[acmsmall,screen, authorversion]{acmart}

\usepackage{amsmath,amsfonts}
\usepackage{algorithm}[H]
\usepackage{array}
\usepackage{subcaption}
\usepackage{textcomp}
\usepackage{stfloats}
\usepackage{url}
\usepackage{verbatim}
\usepackage{graphicx}

\usepackage[noend]{algpseudocode}
\usepackage{xcolor}
\usepackage{xspace}
\usepackage{booktabs} 
\usepackage{multirow}
\usepackage{makecell}
\usepackage{enumitem}
\usepackage{bm}

\usepackage{amssymb} 
\usepackage[capitalise,noabbrev]{cleveref}
\usepackage{etoolbox}

\newcommand{\mapelites}{\textsc{map-elites}\xspace}
\newcommand{\tdthree}{\textsc{td3}\xspace}

\newcommand{\mcx}{\textsc{mcx}\xspace}

\newcommand{\move}{\textsc{move}\xspace}
\newcommand{\dcg}{\textsc{dcg-me}\xspace}
\newcommand{\dcrl}{\textsc{dcrl-me}\xspace}

\newcommand{\moqd}{\textsc{moqd}\xspace}
\newcommand{\moqdlong}{Multi-Objective Quality-Diversity\xspace}
\newcommand{\moea}{\textsc{moea}\xspace}
\newcommand{\qd}{\textsc{qd}\xspace}
\newcommand{\mo}{\textsc{mo}\xspace}
\newcommand{\momdp}{\textsc{momdp}\xspace}
\newcommand{\morl}{\textsc{morl}\xspace}
\newcommand{\rl}{\textsc{rl}\xspace}
\newcommand{\mdp}{\textsc{MDP}\xspace}

\newcommand{\ga}{\textsc{ga}\xspace}
\newcommand{\pg}{\textsc{pg}\xspace}

\newcommand{\initialisation}{// \texttt{Initialisation}\xspace}
\newcommand{\mainloop}{// \texttt{Main loop}\xspace}
\newcommand{\sample}{// \texttt{Sample solutions}\xspace}
\newcommand{\generate}{// \texttt{Generate offspring}\xspace}
\newcommand{\offspringeval}{// \texttt{Evaluate offspring}\xspace}

\newcommand{\archiveadd}{// \texttt{Add to archive}\xspace}
\newcommand{\updateiter}{// \texttt{Update iterations}\xspace}

\newcommand{\replications}{\textsc{20}\xspace}

\newcommand{\anttwo}{\textsc{ant-2}\xspace}
\newcommand{\antthree}{\textsc{ant-3}\xspace}
\newcommand{\halfcheetahtwo}{\textsc{halfcheetah-2}\xspace}
\newcommand{\hoppertwo}{\textsc{hopper-2}\xspace}
\newcommand{\hopperthree}{\textsc{hopper-3}\xspace}
\newcommand{\walkertwo}{\textsc{walker-2}\xspace}

\newcommand{\numpgsteps}{n\xspace}

\newcommand{\moqdscore}{\textsc{moqd-score}\xspace}
\newcommand{\moqdsparsityscore}{\textsc{moqd-sparsity-score}\xspace}
\newcommand{\globalsparsity}{\textsc{global-sparsity}\xspace}
\newcommand{\maxsumscores}{\textsc{maximum sum of scores}\xspace}
\newcommand{\globalhypscore}{\textsc{global-hypervolume}\xspace}
\newcommand{\coverage}{\textsc{coverage}\xspace}

\newcommand{\mome}{\textsc{mome}\xspace}
\newcommand{\pgx}{\textsc{\mome -pgx}\xspace}
\newcommand{\pcclong}{\textbf{M}ulti-\textbf{O}bjective \textbf{M}ap-\textbf{E}lites with \textbf{P}reference-\textbf{C}onditioned Policy-Gradient and \textbf{C}rowding Mechanisms\xspace}
\newcommand{\pcc}{\textsc{\mome -p2c}\xspace}
\newcommand{\pga}{\textsc{pga-me}\xspace}
\newcommand{\spea}{\textsc{spea2}\xspace}
\newcommand{\nsga}{\textsc{nsga-ii}\xspace}
\newcommand{\nsgasmall}{\textsc{nsga-ii-small}\xspace}
\newcommand{\speasmall}{\textsc{spea2-small}\xspace}
\newcommand{\pgmorl}{\textsc{pg-morl}\xspace}

\newcommand{\noactor}{\textsc{no-actor}\xspace}
\newcommand{\nocrowding}{\textsc{no-crowding}\xspace}
\newcommand{\noga}{\textsc{no-ga}\xspace}
\newcommand{\keeppref}{\textsc{keep-pref}\xspace}
\newcommand{\onehot}{\textsc{one-hot}\xspace}

\newcommand{\preference}{\boldsymbol{\omega}}

\newcommand{\code}[0]{\url{https://github.com/adaptive-intelligent-robotics/MOME-P2C.git}}

\begin{document}

\title{Preference-Conditioned Gradient Variations for \moqdlong}

\author{Hannah Janmohamed}
\orcid{https://orcid.org/0000-0001-7997-8455}
\affiliation{%
    \institution{Imperial College London, InstaDeep}
    \city{London}
    \country{UK}
}
\author{Maxence Faldor}
\orcid{https://orcid.org/0000-0003-4743-9494}
\affiliation{%
    \institution{Imperial College London}
    \city{London}
    \country{UK}
}
\author{Thomas Pierrot}
\orcid{https://orcid.org/0000-0002-5227-6194}
\affiliation{%
    \institution{InstaDeep}
    \city{Boston}
    \country{USA}
}
\author{Antoine Cully}
\orcid{https://orcid.org/0000-0002-3190-7073}
\affiliation{%
    \institution{Imperial College London}
    \city{London}
    \country{UK}
}

\begin{abstract}
In a variety of domains, from robotics to finance, Quality-Diversity algorithms have been used to generate collections of both diverse and high-performing solutions.
Multi-Objective Quality-Diversity algorithms have emerged as a promising approach for applying these methods to complex, multi-objective problems.
However, existing methods are limited by their search capabilities. 
For example, Multi-Objective Map-Elites depends on random genetic variations which struggle in high-dimensional search spaces.
Despite efforts to enhance search efficiency with gradient-based mutation operators, existing approaches consider updating solutions to improve on each objective separately rather than achieving desired trade-offs.
In this work, we address this limitation by introducing Multi-Objective Map-Elites with Preference-Conditioned Policy-Gradient and Crowding Mechanisms: a new \moqdlong algorithm that uses preference-conditioned policy-gradient mutations to efficiently discover promising regions of the objective space and crowding mechanisms to promote a uniform distribution of solutions on the non-dominated front.
We evaluate our approach on six robotics locomotion tasks and show that our method outperforms or matches all state-of-the-art \moqdlong methods in all six, including two newly proposed tri-objective tasks.
Importantly, our method also achieves a smoother set of trade-offs, as measured by newly-proposed sparsity-based metrics.

\end{abstract}

\ccsdesc[500]{Theory of computation~Evolutionary algorithms}
\ccsdesc[500]{Applied computing~Multi-criterion optimization and decision-making}
\ccsdesc[300]{Computer systems organization~Evolutionary robotics}

\keywords{Quality-Diversity, Multi-Objective Optimisation, MAP-Elites, Neuroevolution}

\begin{teaserfigure}
\centering
\includegraphics[width =\hsize]{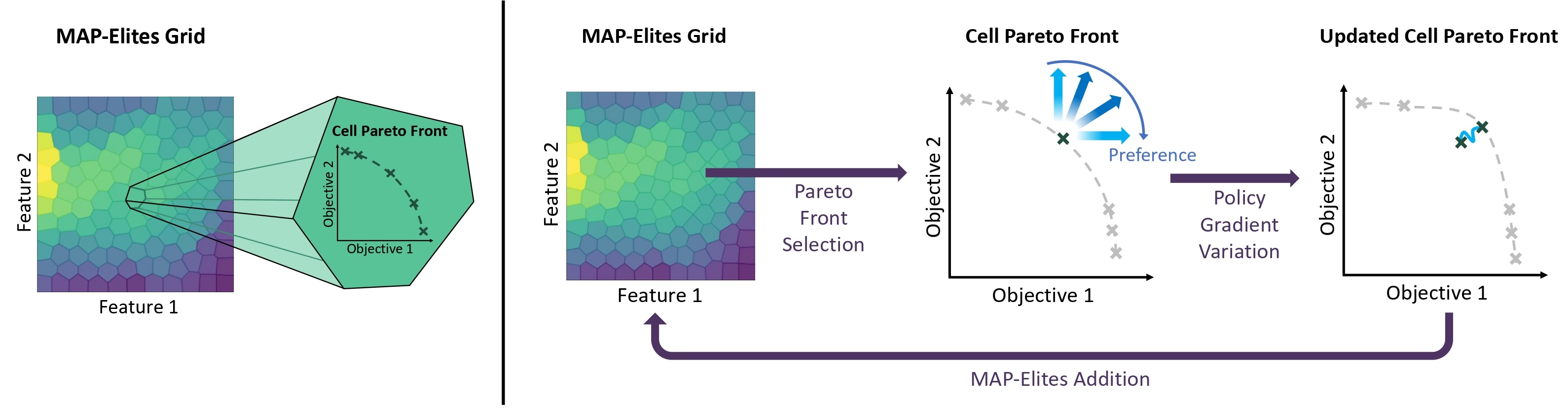}
\caption{\textit{Left. }Multi-Objective MAP-Elites repertoire. The feature space $\mathcal{C} \subset \mathbb{R}^d$ is tessellated into cells $\mathcal{C}_i$. A set of non-dominated solutions is stored in each cell. The aim of \moqd algorithms is to fill each cell with solutions that are Pareto-optimal. \textit{Right.} Overview of preference-conditioned policy gradient in the \pcc algorithm. By conditioning policy-gradients on updates solutions can be improved toward achieving different trade-offs of objectives (illustrated by blue arrows). By contrast, in \pgx, solutions are only updated to improve performance on each objective separately (illustrated by light blue arrows).}
\label{fig:teaser}
\end{teaserfigure}

\maketitle

\section{Introduction}

Over recent years, Deep Reinforcement Learning (RL) has enabled breakthroughs in mastering games~\cite{alphago, dqn} as well as continuous control domains for locomotion~\cite{smith2022walk, cheng2024extreme} and manipulation~\cite{openai2019solving}. These milestones have demonstrated the extraordinary potential of RL algorithms to solve specific problems. However, most approaches return only one highly-specialised solution to a single problem.
In contrast, there is a growing shift in focus towards not just uncovering one single solution that achieves high rewards, but instead many solutions that exhibit different ways of doing so \cite{alphazero_db}.
Within this context, Quality-Diversity (\qd) algorithms \cite{qdunifying} have emerged as one promising approach for tackling this challenge.

In \qd, the primary goal is to produce a variety of high-quality solutions, rather than to focus exclusively on finding the single best one.
One motivation for \qd algorithms is that, finding many solutions can provide availability of alternative, back-up solutions in the event that the highest-performing solution is no longer suitable.
For example, in robotics, generating large collections of solutions has been shown to be helpful for addressing large simulation to reality gaps \cite{nature} and adapting to unforeseen damages \cite{nature, hbr}.
Alternatively, having multiple solutions can simply be used in order to promote innovation in the downstream task. 
In this context, \qd has been used for creating diverse video game levels \cite{videogames,mario_videogames} and generating building designs \cite{mcx}.

Despite the growing traction of \qd, most research in this field has focused on single-objective applications.
However, multi-objective (\mo) problems pervade many real-world domains, including engineering \cite{mo_satellites, mo_windturbines}, finance \cite{mo_finance}, and drug design \cite{mo_molecular} and many state-of-the-art \mo algorithms originate from the Evolutionary Algorithm community \cite{spea, spea2, nsga2, moea/d}.

Recently, Multi-Objective MAP-Elites algorithm (\mome) \cite{mome} marked the first attempt at bridging ideas from \qd and \mo optimisation.
In \moqd, the overarching goal is to identify a broad collection of solutions that exhibit diverse features \textit{and} achieve distinct performances across multiple objectives.
More specifically, given a feature space that is tessellated into cells, the aim is to find a collection of solutions within each cell which offer different trade-offs on each of the objectives (see \Cref{fig:teaser}). 
As an example, consider the task of designing building sites.
Within this context, it may be interesting to find different designs that vary in the number of buildings on the site.
Then for each possible number of buildings, further options can be generated which present different trade-offs of ventilation and noise levels \cite{mcx}.
This approach equips end-users with a spectrum of viable options, thereby broadening their perspective on the array of feasible design possibilities.

The \mome algorithm demonstrated promising results in finding large collections of diverse solutions that balance multiple objectives.
However, \mome predominantly depends on random genetic variations that can cause slow convergence in large search spaces \cite{pga,pga_empirical_analysis,qdpg}.
This renders it less suitable for evolving neural networks with a large number of parameters.

Since the inception of the \mome framework, several related works exploring the domain of \moqd have emerged \cite{mome-pgx,mcx, move}. Among them, \pgx \cite{mome-pgx} builds upon the \mome framework and was shown to achieve state-of-the-art performance on high-dimensional continuous control robotics tasks that can be framed as Markov Decision Processes.
It uses crowding addition and selection mechanisms to encourage an even distribution of solutions on the non-dominated front and employs policy-gradient mutations for each objective function in order to drive the exploration process toward promising regions of the solution space.
However, the \pgx approach is not without its own set of challenges.
Firstly, it employs separate actor-critic networks for each objective function, which can be resource-intensive and may not scale with an increasing number of objectives.
Furthermore, although using policy gradient-based updates helps with exploration in high-dimensional search spaces, the approach in \pgx only considers improving solutions on each objective separately.
However, in the context of multi-objective problems, the goal is often not just to maximise each objective independently but rather to find solutions which offer different trade-offs among them.
In this way, if end users have different \textit{preferences} regarding the relative importance of each objective, they have a range of solutions to choose from.

In this paper, we address the limitations of \pgx by introducing a new \moqd algorithm: \textbf{M}ulti-\textbf{O}bjective \textbf{M}ap-\textbf{E}lites with \textbf{P}reference-\textbf{C}onditioned Policy-Gradient and \textbf{C}rowding Mechanisms (\pcc).
Rather than using a separate actor-critic framework for each objective, \pcc uses a single preference-conditioned actor and a single preference-conditioned critic.
Similar to \pgx, the actor-critic framework in \pcc can be used to provide policy-gradient mutations which offer efficient search space exploration for high-dimensional neural-network policies. 
However, as illustrated in \Cref{fig:teaser}, by conditioning the actor and critic networks on a preference, policy-gradient updates can be used to improve solutions toward achieving a given weighting over the objectives, rather than improve solutions on each objective disjointly.
Moreover, using a single preference-conditioned actor-critic framework rather than one per objective also reduces training costs associated with maintaining the separate actor-critic networks of \pgx.

We show that \pcc outperforms or matches the performance of \pgx across six robotic control \moqd tasks, including newly introduced tri-objective ones (see \Cref{section:tasks}).
\pcc also outperforms \pgx on two newly introduced sparsity-based \moqd metrics (see \Cref{section:metrics}) demonstrating that it is able to attain a smoother set of trade-offs than \pgx.
The code for \pcc is fully containerised and available at \code.



\section{Background}
\subsection{Quality-Diversity}
Quality-Diversity algorithms aim to discover collections of solutions that are both high-performing and diverse~\cite{pugh2016quality}.
Similar to standard optimisation algorithms, a solution $\theta \in \Theta$ is assessed via a fitness function $f: \Theta \to \mathbb{R}$ that reflects its performance on the task.
For example, consider the task of generating an image of a celebrity from a text prompt.
In this case, the fitness of a solution could be the CLIP score \cite{clipscore} which measures the fidelity of an image to its caption that was used to generate it.
However, an additional central component to \qd algorithms, is the concept of the feature function $\Phi: \Theta \to \mathbb{R}^d$ that characterizes solutions in a meaningful
way for the type of diversity desired~\cite{pugh2016quality}.
The feature of a solution $\Phi(\theta_i)$ is a vector that captures some of its notable characteristics, which is then consequently used to quantify its novelty relative to other solutions.
In the image generation example, the feature could be the hair length or age of the subject in the photo \cite{cma-mae}.
In this example, the \qd algorithm would then aim to generate images in which the subject has a diverse range of hair lengths and ages, and which closely obey the given text prompt used to generate it.

One branch of algorithms in the \qd family stems from the \mapelites algorithm~\cite{mapelites}, which has gained prominence for its simplicity and effectiveness.
\mapelites operates by discretising the feature space into a grid-like structure, where each cell $C_i$ of the grid becomes a ``niche" that can be occupied by a solution. 
Tessellating the feature space in this manner creates a systematic method for exploring of different niches within this space \cite{cvt}.
Each iteration of \mapelites first involves selecting solutions from these niches, creating copies of them and mutating these copies to create new candidate solutions.
Then, the fitness and features of the candidate solutions are evaluated, and they are added to the appropriate niches based on their fitness.
If the cell corresponding to the new solution's feature vector is unoccupied, the new solution is added to the cell.
If the cell is occupied, but the evaluated solution has a higher fitness than the current occupant, it is added to the grid.
Otherwise, the solution is discarded.
This process continues for a fixed number of iterations, progressively populating the grid structure with an array of diverse, high-quality solutions.

\mapelites algorithms aim to maximise the total number of occupied cells at the end of the process and the performance of the solutions within each of them. 
Given a search space $\Theta$ and a feature space $\mathcal{C}$ that has been tessellated into $k$ cells $\mathcal{C}_i$, the \mapelites objective, or QD-score~\cite{qdunifying} can be formally expressed as:
\begin{equation}
\max_{\theta\in\Theta} \sum_{i=1}^{k} f(\theta_i), \,\,\text{where} \,\,\forall i, \Phi(\theta_i)\in \mathcal{C}_i\,.
\end{equation}

\subsection {Multi-Objective Optimisation}

Multi-Objective (\mo) optimisation provides an approach for addressing problems that involve the simultaneous consideration of multiple, often conflicting objectives $\textbf{F} = [f_1, \dots, f_m]$.
In \mo problems, objectives often compete with each other, meaning that improving one objective typically comes at the expense of another.
For example, in engineering, improving performance might increase cost, and vice versa.
To navigate this landscape, the concept of Pareto-dominance is commonly employed to establish a preference ordering among solutions.
A solution $\theta_1$ is said to dominate another solution $\theta_2$ if it is equal or superior in at least one objective and not worse in any other \cite{guidetomorl}.
That is, $\theta_1 \succ \theta_2$, if $\forall i: f_i(\theta_1) \geq f_i(\theta_2) \wedge \exists j: f_j(\theta_1) > f_j(\theta_2) $.

Solutions that are not dominated by any other solutions are termed \textit{non-dominated}.
Given a set of candidate solutions $\mathcal{S}$, the non-dominated solutions of this set $\theta_i \in S$ collectively form a \textit{non-dominated front}, which represents the boundary of achievable trade-offs among objectives.
The goal of \mo optimisation is to find an approximation to the \textit{optimal Pareto front}, i.e.,  the non-dominated front over the entire search space $\Theta$.

\begin{figure}[ht!]
\centering
\includegraphics[width=0.4\linewidth]{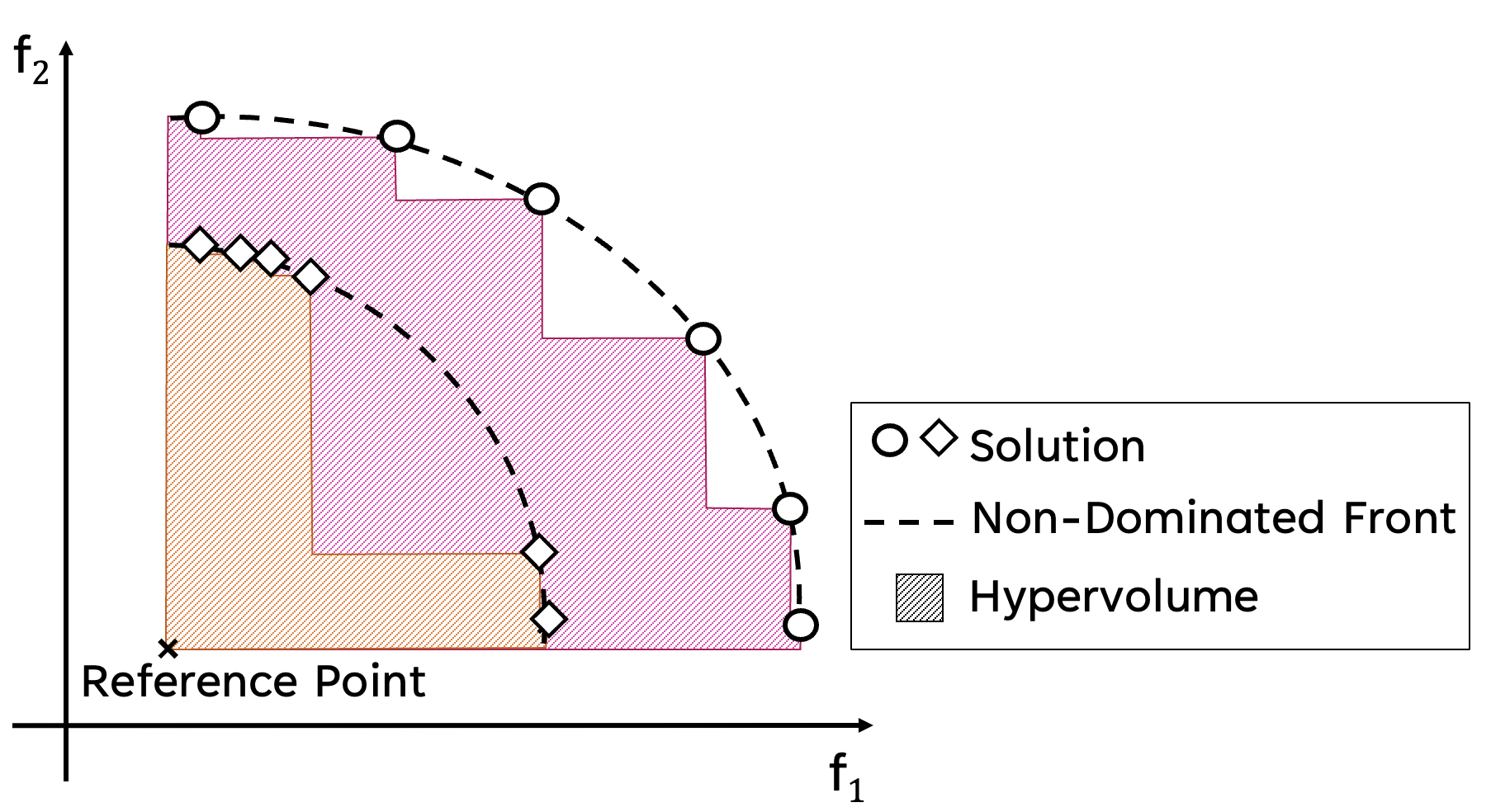}
\caption{
Two sets of solutions that form different non-dominated fronts as approximations of the true Pareto front.
The outer set (marked by circles) achieves a larger hypervolume as it extends further in objective space. Likewise, its sparsity metric is higher, reflecting a more even spread of solutions.
Only the outer set of solutions are considered to be non-dominated when the two sets are combined.}

\label{fig:pfs}
\end{figure}

There are two metrics, the hypervolume and sparsity metric (see \Cref{fig:pfs}), that play pivotal roles in comprehensively assessing the quality and diversity of solutions within a Pareto front approximation \cite{usinghypervolumes, guidetomorl}.
The hypervolume $\Xi$ of a front $\mathcal{P}$, measures the volume of the objective space enclosed by a set of solutions relative to a fixed reference point $r$. 
This metric provides a quantitative measure of the quality and spread of solutions in the objective space and is calculated as \cite{guidetomorl, usinghypervolumes}:
\begin{equation}
    \Xi(\mathcal{P}) = \lambda ({\theta \in \Theta \,|\, \exists \, s \in \mathcal{P}, s \succ x \succ r})
   \label{eqn:hypervolume}
\end{equation}
where $\lambda$ denotes the Lebesgue measure.

While hypervolume is widely used in the multi-objective optimisation literature as a combined measure of convergence and spread, it is also influenced by the reference point and the shape of the front.
For simple linear fronts, hypervolume is maximized by evenly spaced solutions.
But on inverted or irregular fronts it may rank boundary solutions more highly, and its evaluation can vary significantly with the choice of reference point, sometimes leading to counter-intuitive comparisons \cite{hypervolumesurvey, referencepoints}.
We therefore acknowledge these limitations, but we follow prior work in Multi-Objective Quality-Diversity in adopting hypervolume as a standard comparative metric \cite{mome, mome-pgx}.

Sparsity, by contrast, provides complementary information regarding the distribution and evenness of these solutions.
It is calculated by evaluating the average nearest neighbour distance among solutions on the front, given by \cite{guidetomorl}:
\begin{equation}
    S(\mathcal{P}) = \frac{1}{|\mathcal{P}| - 1} \sum_{j=1}^m \sum_{i=1}^{|\mathcal{P}| - 1} (\Tilde{P_j}(i) - \Tilde{P_j}(i+1))^2
   \label{eqn:sparsity}
\end{equation}
where $\Tilde{P_j}(i)$ denotes the $i$-th solution of the list of solutions on the front $\mathcal{P}$, sorted according to the $j$-th objective and $|\mathcal{P}|$ denotes the number of solutions on the front.
To ensure that the sparsity is not skewed due to different scales of each of the objectives, the objective functions must be normalised prior to calculating it.
A low-sparsity metric indicates that solutions are well-dispersed through the objective space, highlighting the algorithm's ability to provide diverse trade-off solutions.
In contrast, a high-sparsity metric suggests that solutions are clustered in specific regions, potentially indicating that the algorithm struggles to explore and represent the full range of possible trade-offs.

\subsection {Multi-Objective Quality-Diversity Algorithms}

Multi-Objective Quality-Diversity (\moqd) combines the goals of QD and MO optimisation. 
Specifically, the goal of \moqd is to return the non-dominated front of solutions in each cell of the feature space with maximum hypervolume, $\mathcal{P}(\mathcal{C}_i)$ \cite{mome}. This \moqd goal can be mathematically formulated as:
\begin{equation}
\max_{\theta\in\Theta} \sum_{i=1}^{k} \Xi(\mathcal{P}_i), \,\,\text{where} \,\,\forall i, \mathcal{P}_i = \mathcal{P}({\theta|\Phi(\theta)\in \mathcal{C}_i})
\label{eqn:moqd}
\end{equation}
\mome \cite{mome} was the first \moqd algorithm that aimed to achieve this \moqd goal.
To achieve this, \mome maintains a non-dominated front in each cell of a MAP-Elites grid.
At each iteration, a cell is uniformly selected and then a solution from the corresponding non-dominated front is uniformly selected.
Then, the algorithm follows a standard \mapelites procedure: the solution undergoes genetic variation and is evaluated.
The evaluated solution is added back to the grid if it lies on the non-dominated front of the cell corresponding to its feature vector.

We note that \moqd differs from related areas such as multi-modal multi-objective optimisation \cite{li2023multimodal, liang2016multimodal}.  
The purpose of multi-modal multi-objective optimisation is to provide a set of equivalent Pareto-optimal outcomes, where each outcome achieves nearly the same balance of objectives but corresponds to a different solution representation, giving end-users flexibility in practice.
For example, finding two different circuit layouts that yield the same power–efficiency trade-off, yet offer engineers different implementation choices.
In other words, in multi-modal multi-objective optimisation, the goal is to identify multiple Pareto-optimal solutions that are \emph{equivalent in objective space} but differ in decision space.  
By contrast, \moqd explicitly introduces a feature space $\Phi(\theta)$ that structures the search according to user-defined features.
The aim is then to return a Pareto front approximation \emph{within each feature cell}, rather than globally across all solutions.  
This distinction means that while multi-modal multi-objective optimisation emphasises diversity of solutions that achieve similar trade-offs, \moqd emphasises diversity across features and simultaneously seeks high-performing trade-offs within each feature category.

\subsection{Problem Formulation}
\def \mathbi#1{\textbf{\em#1}}
In this work, we consider an agent sequentially interacting with an environment for an episode of length $T$, modelled as a Multi-Objective Markov Decision Process (\momdp), defined by $\langle \mathcal{S}, \mathcal{A}, \mathcal{P}, \mathbi{R}, \Omega \rangle$.
At each discrete time step $t$, the agent observes the current state $s_t \in \mathcal{S}$ and takes an action $a_t \in \mathcal{A}$ by following a policy $\pi_\theta$ parameterized by $\theta$. Consequently, the agent transitions to a new state sampled from the dynamics probability distribution $s_{t+1} \sim p(s_{t+1} | s_t, a_t)$. The agent also receives a reward vector $\mathbf{r}_t = [r_1(s_t, a_t), \dots, r_m(s_t, a_t)]$, where each reward function $r_i: \mathcal{S} \times \mathcal{A} \to \mathbb{R}$ defines an objective.
The multi-objective fitness of a policy $\pi$ is defined as a vector \textbf{F}$(\pi) = [f_1(\pi), ..., f_m(\pi)]$. Here, each $f_i$ represents the expected discounted sum of rewards, calculated as $f_i = \mathbb{E}_\pi \left[ \sum_t \gamma^t r_i(s_t, a_t) \right]$ for a given reward function $r_i$. The discount rate $\gamma \in [0, 1]$ controls the relative weighting of immediate and long-term rewards.

\subsection{Reinforcement Learning}\label{section:td3}
In the single-objective case ($m = 1$), the \momdp collapses into a simple Markov Decision Process (\mdp) with scalar rewards, where the goal is to find a policy $\pi$ that maximises the expected discounted sum of rewards or return, $F(\pi) = \mathbb{E}_{\pi}\left[ \sum_t \gamma^t r(s_t, a_t) \right]$.
Numerous Reinforcement Learning (\rl) methods have been developed to address the challenge of finding policies that optimize this cumulative reward.
One particularly relevant approach is the Twin Delayed Deep Deterministic Policy Gradient algorithm (\tdthree)\cite{td3}.

The \tdthree algorithm belongs to the broader family of actor-critic \rl techniques \cite{actor-critic-survey}, which involve two key components: an actor network and a critic network.
The actor network is a policy parameterised by $\phi$, denoted $\pi_\phi$ that is used to interact with the environment. The transitions $(s_t, a_t, r_t, s_{t+1})$ coming from the interactions with the environment are stored in a replay buffer $\mathcal{B}$ and used to train the actor and the critic.
The critic network is an action-value function parameterised by $\psi$, denoted $Q_\psi$ that evaluates the quality of the actor's actions and helps the agent learn to improve its decisions over time.
The critic estimates the expected return obtained when starting from state $s$, taking action $a$ and following policy $\pi$ thereafter, $Q_\psi(s, a) = \mathbb{E}_\pi[\sum_t \gamma^t r(s_t, a_t) | s_0 = s, a_0 = a]$.

The \tdthree algorithm, uses a pair of critic networks $Q_{\psi_1}, Q_{\psi_2}$, rather than a single critic network in order to reduce overestimation bias and mitigate bootstrapping errors.
These networks are trained using samples $(s_t, a_t, r_t, s_{t+1})$ from the replay buffer and then regression to the same target:
\begin{equation}
    y = r(s_t, a_t) + \gamma \min_{i=1,2} Q_{ \psi_i}(s_{t+1}, \pi_{\phi'}(s_{t+1}) + \epsilon)
 \label{eqn:qtarget}
\end{equation}
where $Q_{\psi_1'}, Q_{\psi_2'}$ and $\pi_{\phi'}$ are target networks that are used in order to increase the stability of the training and $\epsilon$ is sampled Gaussian noise to improve exploration and smoothing of the actor policy.
The actor network is updated to choose actions which lead to higher estimated value according to the first critic network $Q_{\psi_1}$.
This is achieved via a policy gradient (\pg) update:
\begin{equation}\label{eqn:actor_policygradient}
    \nabla _{\phi} J(\pi_\phi) = \mathbb{E}\big[\,\nabla _{\phi}\,\pi _{\phi}(s)\,\nabla _a Q_{\psi_1} (s, a)\,|\,_{a=\pi _{\phi}(s)} \big]
\end{equation} 
These actor \pg updates are executed less frequently than the critic network training in order to enhance training stability.

\section{Related Works}
\subsection{Multi-Objective Evolutionary Algorithms}
\label{section:moeas}
Multi-Objective Evolutionary Algorithms (\moea) evolve a population of potential solutions iteratively over several generations to identify an optimal set of solutions that balance conflicting objectives.
At each iteration, solutions are selected from the population and undergo genetic variation (through crossover and mutation operators) and are then added back to the population.
Different MOEAs can vary in terms of their specific selection strategies, crossover and mutation operators, population management techniques, and how they maintain diversity in the population \cite{moea_survey}.

Non-dominated Sorting Genetic Algorithm II (\nsga) \cite{nsga2} and Strength Pareto Evolutionary Algorithm 2 (\spea) \cite{spea2} both use non-uniform selection mechanisms to guide the optimisation process.
Both methods select solutions that are higher performing and occupy less dense regions of the objective space with higher probability.
This guides the population towards higher-performing fronts, while simultaneously ensuring solutions are well-distributed across the front.

Our method, \pcc has synergies with many methods from \moea literature including non-uniform selection and addition mechanisms (see \Cref{section:crowding}) and we refer the interested reading to a comprehensive survey of \moea algorithms for more details \cite{moea_survey}.
However, we emphasise that \moea algorithms treat all objectives as quantities to maximise, seeking trade-offs among them.
However, in \moqd algorithms, the goal is not to \textit{maximise} feature values but to \textit{cover} the feature space with high-performing solutions.
In other words, \moqd algorithms seek solutions which are diverse in the \textit{feature}-space, and then to maximise objectives for each feature.
Moreover, our method differs from traditional \moea approaches in two significant aspects.
First, it employs a \mapelites grid to explicitly maintain solutions that are diverse in feature space while optimising over objectives.
Second, it incorporates techniques from reinforcement learning to form gradient-based mutations which help to overcome the limited search power of traditional \ga variations for high-dimensional search spaces \cite{pga}.

\subsection{Multi-Objective Reinforcement Learning}
In multi-objective reinforcement learning (\morl) the expected sum of rewards is a vector $\boldsymbol{J}(\pi) = \mathbb{E}_{\pi}[\sum_t \boldsymbol{r_t}]$. Consequently, there is not a straightforward notion of a reward maximising agent.
Single-policy \morl approaches focus on discovering a single policy that achieves a desired trade-off of objectives. Often, this is achieved by employing a scalarization function which transforms the performance on various objectives into a single scalar utility value.
For example, many approaches aim to find a policy $\pi$ that maximises the expected weighted sum of rewards,
\begin{equation}\label{eqn:weighted_expected_returns}
\begin{split}
J(\pi, \preference)=\mathbb{E}_{\pi}\Big[\sum_t \preference^\intercal \boldsymbol{r}_t\Big]=\preference^\intercal \mathbb{E}_{\pi}\Big[\sum_t \boldsymbol{r}_t\Big] =\preference^\intercal \boldsymbol{J}(\pi)
\end{split}
\end{equation}
Here, $\preference$ is referred to as a \textit{preference}, with $\sum_i \omega_i = 1$.
The preference quantifies the relative importance of each of the objective functions and, when the preference is fixed, we can collapse the \momdp into a single-objective setting that can be optimised with well-established \rl approaches.

In single-policy approaches, the challenge arises in determining the preference vector beforehand, as it may prove to be a complex task or may vary among different users \cite{guidetomorl}.
Instead, it may be useful to find solutions which are optimal for different preference values so that the user can examine the range of possible solutions that is on offer and then assign their preferences retrospectively \cite{guidetomorl}.
With this perspective in mind, multi-policy \morl methods aim to find a set of policies that excel across a range of different preferences \cite{dols, prediction_guided_morl}.
Often, each policy in the set is trained using preference-conditioned policy-gradient derived from a multi-objective, preference-conditioned action-value function \cite{dols, prediction_guided_morl, generalised-morl}.


Some methods straddle the line between single-policy and multi-policy \morl by seeking a single preference-conditioned policy that can maximise the weighted sum of expected returns (\cref{eqn:weighted_expected_returns}) for any given preference \cite{parisi2016multi, dynamicweights, generalised-morl, scalingparetodecisions}.
This approach offers advantages such as reduced storage costs and rapid adaptability \cite{dynamicweights}.
However, while having preference-conditioned policy approaches might be cheaper and more flexible, these methods have been observed to achieve worse performance on the objective functions for any given preference than having specialised policies \cite{prediction_guided_morl}.

Our work combines elements of both preference-conditioned and multi-policy approaches.
Our actor-critic networks are preference-conditioned and are separate from the \mapelites grid.
Hence this preference-conditioned actor can be seen as a generalist policy.
However, within each cell of the \mapelites grid, we store specialised, non-conditioned policies hence taking a multi-policy approach.
To the best of our knowledge, there is no prior research in multi-objective reinforcement learning (\morl) that actively seeks to diversify behaviours in this manner.



\subsection{Gradients in Quality-Diversity}
\label{section:qd}
\qd algorithms belong to the wider class of Genetic Algorithms (\ga), which broadly adhere to a common structure of selection, variation and addition to a population of solutions.
While these methods have been observed to be highly-effective black box methods, one key limitation is their lack of scalability to high-dimensional search spaces.
In tasks in which solutions are the parameters of a neural network, the search space can be thousands of dimensions and thus traditional \ga variation operators do not provide sufficient exploration power. 
To address this, many works in single-objective \qd leverage the search power of gradient-based methods in high-dimensional search spaces \cite{pga, pga_empirical_analysis, qdpg, dqd, dqdformdps, dcg}.
The pioneer of these methods, Policy-gradient assisted \mapelites (\pga) \cite{pga}, combines the \tdthree algorithm with the \mapelites algorithms to apply \qd to high-dimensional robotics control tasks.
In particular, during the evaluation of solutions in \pga, environment transitions are stored and used to train actor and critic networks, using the training procedures explained in \Cref{section:td3}.
Then, \pga follows a normal \mapelites loop except in each iteration, half of the solutions are mutated via \ga variations and the other half are mutated via policy-gradient (\pg) updates.

Since \pga, several other \qd algorithms with gradient-based variation operators have been proposed.
Some of these are tailored to consider different task settings which have differentiable objective and feature functions \cite{dqd} or  discrete action spaces \cite{discrete_illumination}.
Other methods use policy gradient updates to improve both the fitness \textit{and} diversity of solutions \cite{qdpg, dqd, dqdformdps}.
A particular method of note is \dcg~\cite{dcg,dcrl} which uses policy-gradient variations conditioned on features of solutions.
Similar to \pcc, the motivation for this method is to provide more nuanced gradient information. 
Conditioning the policy-gradient on the feature value of a solution provides a way to update the solution toward higher performance, \textit{given} that it has a certain behaviour.
However, this method only considered mono-objective problems.
Other than \pgx (see \Cref{section:moqd}) we are unaware of gradient-based \qd methods applied multi-objective problems.


\subsection{Multi-Objective Quality-Diversity Algorithms}
\label{section:moqd}
Recently, policy gradient variations, inspired by single-objective methods, have played a pivotal role in shaping the development of techniques in \moqd. 
Notably, while \mome (see \Cref{section:moqd}) is a simple and effective \moqd approach, it relies on \ga policy-gradient mutations as an exploration mechanism which makes it inefficient in high-dimensional search spaces.
To overcome this challenge, Multi-Objective MAP-Elites with Policy-Gradient Assistance and Crowding-based Exploration (\pgx) \cite{mome-pgx} was recently introduced as an effort to improve the performance and data-efficiency of \mome in tasks that can be framed as a \momdp.
\pgx maintains an actor and critic network for each objective function separately and uses policy gradient mutation operators in order to drive better exploration in the solution search space.
\pgx also uses crowding-based selection and addition mechanisms to bias exploration in sparse regions of the non-dominated front and to maintain a uniform distribution of solutions on the front.
\pgx was shown to outperform \mome and other baselines across a suite of multi-objective robotics tasks involving high-dimensional neural network policies.
However, since each actor-critic network pair learns about each of the objective separately, the \pg variations in \pgx may only provide disjoint gradient information about each of the objectives, and fail to capture nuanced trade-offs.

To the best of our knowledge \mome and \pgx are the only existing \moqd algorithms to date. However, we also note of two particularly relevant approaches which have synergies with the \moqd setting.
Multi-Criteria Exploration (\mcx) \cite{mcx} which uses a tournament ranking strategy to condense a solution's score across multiple objectives into a single value, and then uses a standard MAP-Elites strategy.
Similarly, Many-objective Optimisation via Voting for Elites (\move) \cite{move} uses a MAP-Elites grid to find solutions which are high-performing on many-objective problems.
In this method, each cell of the grid represents a different subset of objectives and a solution replaces the existing solution in the cell if it is better on at least half of the objectives for the cell.
While both \mcx and \move consider the simultaneous maximisation of many objectives, they both aim to find one solution per cell in the \mapelites grid rather than non-dominated fronts for different features. Therefore, we consider their goals to be fundamentally different from the \moqd goal defined in \Cref{eqn:moqd}.

Additionally, several works have used multi-objective mechanisms to enhance single-objective \qd. These works consider quality (fitness) and diversity directly as optimisation objectives, rather than task-based objectives. 
For instance, novelty search with local competition \cite{nslc} balances novelty and performance through a multi-objective selection scheme, indicator-based methods \cite{neumann2019evolutionary} employ measures such as hypervolume to guide the spread of solutions, and Pareto-based selection approaches \cite{multiobjective-selection-qd, do2024evolutionary} explicitly use diversity criteria in archive updates.
In all cases, diversity acts as a meta-objective that shapes the search, but the final output remains a single repertoire of diverse, high-quality solutions, in contrast to \moqd which seeks Pareto fronts of task-based trade-offs within each feature.

\section{MOME-P2C}
\begin{figure}[ht!]
\centering
\includegraphics[width=0.6\linewidth]{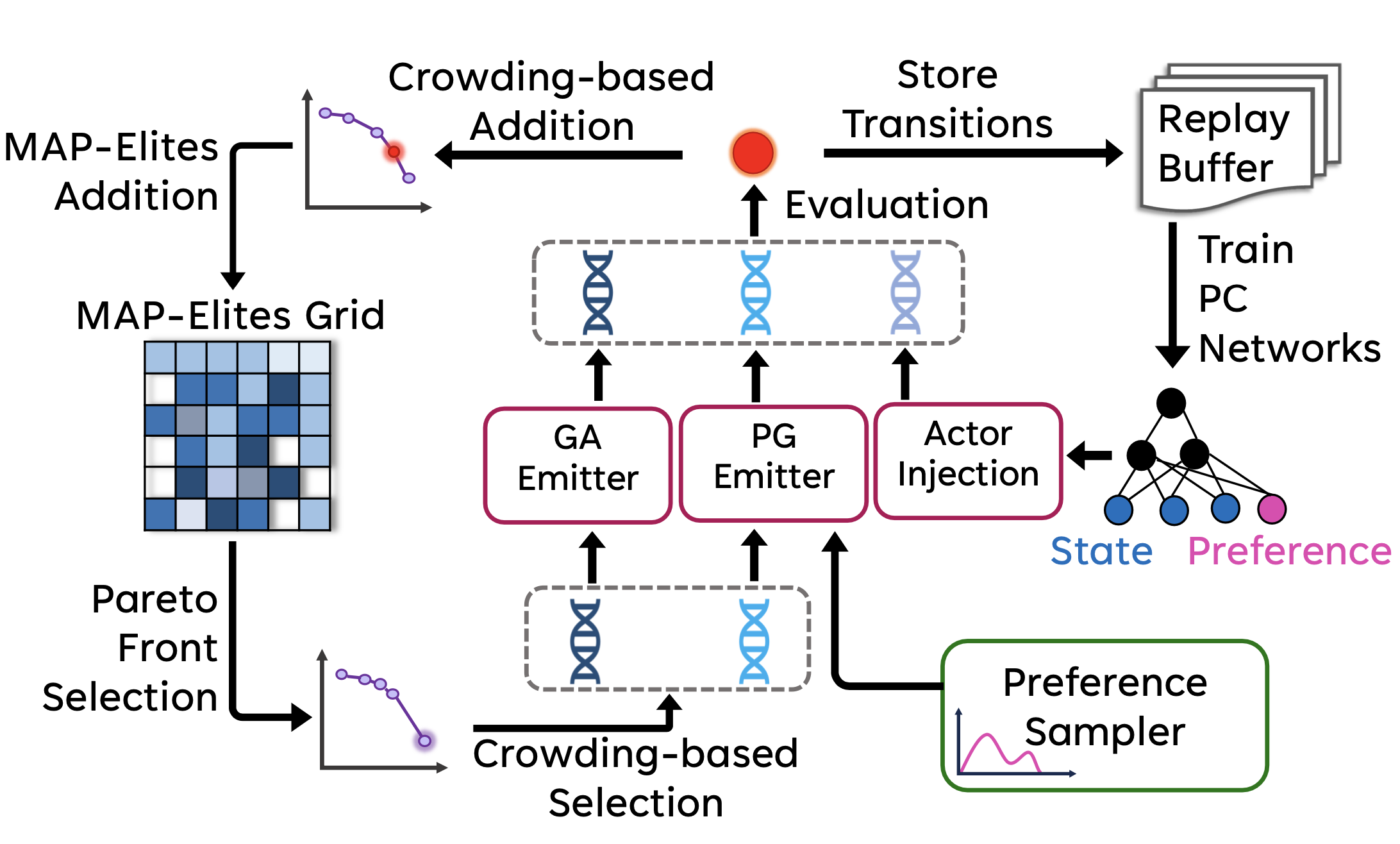}
\caption{Overview of \pcc algorithm. Non-dominated Fronts are stored in each cell of a \mapelites grid. At each iteration, a batch of solutions are selected, undergo variation and are added back to the grid based on their performance and crowding-distances. As solutions are evaluated, environment transitions are gathered in a replay buffer and used to train preference-conditioned networks. These networks are used with a preference sampler to perform preference-conditioned \pg updates.}
\label{fig:method}
\end{figure}

In this section, we introduce \pcclong (\pcc), a new \moqd algorithm that learns a single, preference-conditioned actor-critic framework to provide policy-gradient variations in tasks that can be framed as \mdp.
The algorithm inherits the core framework of existing \moqd methods, which involves maintaining a non-dominated front within each feature cell of a \mapelites grid and follows a \mapelites loop of selection, variation, and addition for a given budget.
Building on the approach of \pgx, our method not only employs traditional genetic variation operators but also integrates policy gradient mutations that improve sample-efficiency, particularly in high-dimensional search spaces.
Similar to \pgx, \pcc adopts crowding-based selection, which strategically directs exploration towards less explored areas of the search space and also utilizes crowding-based addition mechanisms to promote a continuous distribution of solutions along the non-dominated front.
Distinct from \pgx, which operates with a separate actor-critic framework for each objective function, \pcc innovates by employing a singular, preference-conditioned actor-critic. This design streamlines preference-conditioned policy gradient variation updates to genotypes, reducing the memory requirements of the algorithm and making it more scalable to problems with a higher number of objectives (a numerical comparison is shown in \Cref{app:pghyperparams}).
Furthermore, \pcc leverages the preference-conditioned actor by injecting it into the main population.
A visual representation of the algorithm is depicted in \Cref{fig:method}, and the accompanying pseudo-code is provided in \Cref{alg:pseudo}. Detailed descriptions of each component of \pcc are available in the following sections.

\begin{algorithm}
\caption{\pcc pseudo-code}\label{alg:pseudo}
\begin{algorithmic}

\State{\textbf{Input:}}
\begin{itemize}
    \item \mome archive $\mathcal{A}$ and total number of iterations $N$
    \item PG batch size $b_p$, GA batch size $b_g$ (with $b = b_p+b_g$) and actor injection batch size $b_a$
\end{itemize}

\\
\State{\initialisation}
\State{Initialise archive $\mathcal{A}$ with random solutions $\theta_k$}
\State{Initialise replay buffer $\mathcal{B}$ with transitions from $\theta_k$}
\State{Initialise actor and critic networks $\pi_{\phi}$, $\boldsymbol{Q_{\psi_1}}$, $\boldsymbol{Q_{\psi_2}}$}
\\

\State{\mainloop}
\For{$i = 1 \to N $}
    
    \State{\sample}
    \State{$\theta_1, ..., \theta_{b} \gets \text{crowding\_selection}(\mathcal{A})$}
    
    \\
    
    \State{\generate}
    \State{$\preference_1, ..., \preference_{b_{p}} \gets \text{preference\_sampler}({\theta_{1}, ..., \theta_{b_{p}}})$}
    
    \State{$\Tilde{\theta_{1}}, ..., \Tilde{\theta_{b_{p}}} \gets \text{pg\_variation}(\theta_{1}, ..., \theta_{b_{p}}, \preference_1, ... ,\preference_{b_{p}} $)}
    
    \State{$\Tilde{\theta_{b_p+1}}, ..., \Tilde{\theta_{b}} \gets \text{ga\_variation}(\theta_{b_p +1}, ..., \theta_{b}$)}

    \State{$\preference_{b_{p}+1}, ... , \preference_{b_{p} + b_{a}} \gets \text{actor\_sampler}$} 
    \State{$\Tilde{\theta_{b+1}},...,\Tilde{\theta_{b+b_{a}}}\gets \text{actor\_inject}(\pi_\phi, \preference_{b_{p}+1}, ... , \preference_{b_{p}+b_{a}} $)}
    
    \\
    \State{\offspringeval}
    \State{$(f_1,...,f_m, d, \text{transitions})\gets \text{evaluate}(\pi_{\Tilde{\theta_{1}}},...,\pi_{\Tilde{\theta_{b+b_{a}}}})$}
    \State{$\mathcal{B}\gets \text{insert}(\text{transitions})$}
    \State{$\pi_{\phi}, Q_{\psi_1}, Q_{\psi_2} \gets \text{train\_networks}(\mathcal{B}, \pi_{\phi}, \boldsymbol{Q_{\psi_1}}, \boldsymbol{Q_{\psi_2}})$}

    \\
    \State{\archiveadd}
    \State{$\mathcal{A}\gets \text{crowding\_addition}(\Tilde{\theta_{b+1}},...,\Tilde{\theta_{b+b_{a}}})$}
    \\
    \State{\updateiter}
    \State{$i \gets i + 1$}
    \\

\EndFor
\Return{ $\mathcal{A}$}
\end{algorithmic}
\end{algorithm}
\subsection{Crowding-based Selection and Addition}\label{section:crowding}

In \pcc, following \pgx \cite{mome-pgx}, we choose to use biased selection and addition mechanisms.
In particular, when selecting parent solutions from the grid, we first select a cell with uniform probability and then select an individual from the cell's non-dominated front with probability proportional to its crowding distance. 
As defined in \nsga \cite{nsga2}, the crowding distance of a solution is defined as the average Manhattan distance between itself and its nearest neighbours on either side of itself in objective space. 
In \pgx, it was shown that biasing solutions in this manner provides an effective method for guiding the optimisation process toward under-explored regions of the solution space.

Similarly, we also use a crowding-informed addition mechanisms to replace solutions on the non-dominated front.
It is important to note that all \moqd methods  we consider use a fixed maximum size for the non-dominated front of each cell.
This is done in order to exploit the parallelism capabilities of recent hardware advances \cite{qdax, qdaxrepo} and consequently affords many thousands of evaluations in a short period of time.
However, if a solution is added to a non-dominated front that is at already maximum capacity, another solution must also necessarily be removed.
In \pcc, following from \pgx, we remove the solution with the minimum crowding distance in order to increase the sparsity of solutions on the front.


 
Further details regarding the crowding-based mechanisms can be found in the \pgx paper \cite{mome-pgx}.  
While we acknowledge that crowding distance may lose effectiveness in many-objective settings \cite{zheng2024truthful, koppen2007substitute}, we adopted it here as a simple and efficient choice consistent with \pgx, and note that it still yields strong results (we include an ablation of the crowding mechanisms in our ablation study \cref{sec:ablation_results}).  
Nevertheless, we note that exploring alternative approaches for preserving diversity of the non-dominated front with a higher number of objectives remains an interesting avenue for future work.  

\subsection{Preference-Conditioned Actor-Critic}
\label{section:pc-actor-critic}
In \pgx, a separate actor-critic framework was used to find a policy $\pi$ that marginally maximised the expected sum of rewards $\boldsymbol{J}^i(\pi) = \mathbb{E}_{\pi}[\sum_t \boldsymbol{r_t}^i]$ for each objective $i=1, ..., m$.
However, in \pcc, we do not require a separate actor-critic framework for each objective function. 
Instead, we use a \textit{single} actor-critic framework that aims to find a single actor policy to maximise $J(\pi, \preference)=\mathbb{E}_{\pi}[\sum_t \preference^\intercal\boldsymbol{r}_t]$ for any given preference $\preference$.

Accordingly, we modify the actor network $\pi_\phi(s)$ to be a conditioned on a preference $\pi_\phi(s|\preference)$. 
By doing so, the actor network now aims to predict the best action to take from state $s_t$ \textit{given} that its preference over objectives is $\preference$.
In practice, this means that the actor takes its current state $s_t$ concatenated with a preference-vector $\preference$ as input, and outputs an action $a_t$. 

Training a preference-conditioned actor requires a corresponding preference-conditioned critic that evaluates the performance of the actor based on the actor's preference.
In this setting, we take corresponding preference-conditioned action-value function $Q^\pi(s, a | \preference)$ to be:
\begin{equation}\label{eqn:moqvalue}
\begin{split}
    Q^\pi(s, a | \preference) &= \mathbb{E}_{\pi(\cdot|\preference)}\left[\sum_{t=0}^{T} \gamma^t \preference^\intercal \boldsymbol{r}(s_t, a_t) \mid s_0 = s, a_0 = a\right]\\
    &=\preference^\intercal \mathbb{E}_{\pi(\cdot|\preference)}\left[\sum_{t=0}^{T} \gamma^t \boldsymbol{r}(s_t, a_t) \mid s_0 = s, a_0 = a\right]\\
    &=\preference^\intercal \boldsymbol{Q}^{\pi}(s, a)
\end{split}
\end{equation}
Here, $\boldsymbol{Q}^\pi(s, a | \preference)$ denotes the preference-conditioned vectorised action-value function.
\Cref{eqn:moqvalue} demonstrates that the we can estimate the preference-conditioned action-value function by training a critic $\boldsymbol{Q}_\psi(s, a | \preference) \to \mathbb{R}^m$ to predict the vectorised action-value function and then weighting its output by the preference. 
To train this critic network, we modify the target \tdthree algorithm given in \cref{eqn:qtarget} to be:
\begin{equation}
    y = \preference^\intercal \boldsymbol{r}(s_t, a_t) + \gamma \min_{i=1,2} \preference^\intercal \boldsymbol{Q}_{ \psi_i}(s_{t+1}, \pi_{\phi'}(s_{t+1}|\preference) + \epsilon)
 \label{eqn:mo_target}
\end{equation}
which we estimate from minibatches of environment transitions $(s_t, a_t, \boldsymbol{r_t}, s_{t+1})$ stored in the replay buffer $\mathcal{B}$.

In order to train the preference-conditioned actor, we use a preference-conditioned version of the policy gradient from \Cref{eqn:actor_policygradient}:
\begin{equation}
\label{eqn:mo_actor_pg}
\begin{split}
    \nabla _{\phi} J(\phi, \preference) &=
    \preference^\intercal  \mathbb{E}\big[\,\nabla _{\phi}\,\pi _{\phi}(s|\preference)\,\nabla _a \boldsymbol{Q_\psi} (s, a|\preference)\,|\,_{a=\pi _{\phi}(s|\preference)} \big]\\
    &= \mathbb{E}\big[\,\nabla _{\phi}\,\pi _{\phi}(s|\preference)\,\nabla _a \preference^\intercal\boldsymbol{Q_\psi} (s, a|\preference)\,|\,_{a=\pi _{\phi}(s|\preference)} \big]
\end{split}
\end{equation} 
The updates of the actor and critic networks, given by \Cref{eqn:mo_target} and \Cref{eqn:mo_actor_pg}, depend on the value of the preference $\preference$.
In \pcc, for each sampled transition, we uniformly sample a preference and use this to form a single policy gradient update.
Since the preference vector assumes that each of the objectives are scaled equally, we normalise the reward values using a running mean and variance throughout the algorithm. 
Solutions are stored and added to the archive based on unnormalised fitnesses for interpretability and comparison to other methods.

\subsection{Preference-Conditioned Policy Gradient Variation}
Given the preference-conditioned actor-critic framework described in \Cref{section:pc-actor-critic}, we can form preference-conditioned \pg variations on solutions in the archive.
In \pcc at each iteration, we select $b_p$ solutions from the archive and perform \numpgsteps preference-conditioned policy gradient steps. For a given solution $\theta$ which parametrises a policy, $\pi_{\theta(s)}$ we update the parameters via:
\begin{equation}\label{eqn:pc-pg-update}
    \nabla _{\theta} J(\theta, \preference) = \mathbb{E}\big[\,\nabla _{\theta}\,\pi _{\theta}(s)\,\nabla _a \mathbf{Q_{\psi_1}} (s, a | \preference)\,|\,_{a=\pi _{\theta}(s)} \big]
\end{equation} 
The \pg update given by \Cref{eqn:pc-pg-update} depends on a preference vector $\preference$. 
However, it is not straightforward to determine the best strategy for choosing the value of this vector.
In this work, we use the term ``\pg preference sampler" to refer to the strategy we use for determining the preference that the \pg variation is conditioned on (illustrated in \Cref{fig:method}).
In \pcc, we choose the \pg preference sampler to simply be a random uniform sampler as we found this to be a simple, yet effective strategy.
We examine other choices for the \pg preference sampler in our ablation study (\Cref{section:ablation_results}).

\subsection{Actor Injection}

In \pga, \pgx and other gradient-based \qd methods, the actor policy has the same shape as the policies stored in the \mapelites grid and so can be regularly injected into the main offspring batch as a genotype, with no additional cost to the main algorithm. 
However, in \pcc, the policies in the \mapelites grid only take the current state $s_t$ as input, whereas the preference-conditioned actor takes the state concatenated with a preference $[s_t, \preference]$ as input.
Therefore, the actor has a different architecture to the policies so cannot be added to the repertoire. In this work, we take a similar approach to the one taken by \dcrl to inject the conditioned actor within the population~\cite{dcrl}.

Given the weights $W \in \mathbb{R}^{n \times (|\mathcal{S}|+m)}$ and bias $B\in\mathbb{R}^{n}$ of the first layer of the actor network, we note if we fix the value of $\preference$ we can express the value of the $n$ neurons in the first layer as:
\begin{equation}
\begin{aligned}
W\begin{bmatrix}s_t\\ \preference\end{bmatrix} + B &= \begin{bmatrix}W_s & W_{\preference}\end{bmatrix} \begin{bmatrix}s_t\\ \preference\end{bmatrix} + B\\
&= W_s s_t + (W_{\preference} \preference + B)
\end{aligned}
\end{equation}
In other words, if the input preference $\preference$ to the actor is fixed, we can reshape the preference-conditioned actor network to be the same shape as the policies in the \mapelites grid by absorbing the weights corresponding to the preference input into the bias term of the first layer.
%
%
This method provides a cheap approach to use the preference-conditioned actor network to generate offspring which have the same shape as other policies in the grid.
To take advantage of this in \pcc, at each iteration we sample $n_a$ preferences from an ``actor preference sampler" (see \Cref{fig:method}) and use them to reshape the actor network into $n_a$ new policies.
In this work, we choose the actor preference sampler to generate $n_a - m$ uniformly sampled preference vectors for exploration, and $m$ one-hot vectors (with a one at the index for each of the objectives) to ensure that fitness in each of the objectives is always pushed.

\section{Experimental Setup}
In this section, we describe the evaluation tasks, baselines and metrics we use to evaluate our approach.
We introduce two new tri-objective \moqd tasks, \antthree and \hopperthree, which allow us to evaluate the capabilities of different \moqd approaches to scale to a larger number of objectives.
We also introduce two new \moqd metrics (\moqdsparsityscore and \globalsparsity) which we argue are important ways to assess whether \moqd algorithms are able to achieve smooth sets of trade-offs. These new tasks and metrics form two key contributions of this work.
\begin{table*}[ht!]
\centering
  \caption{Summary of evaluation tasks.}
  \label{tab:tasks}
  \scalebox{0.7}{
  \begin{tabular}{ c | p{2.4cm} | p{2.4cm} | p{2.4cm} | p{2.4cm} | p{2.4cm} | p{2.4cm} }
  
    
    & \multicolumn{2}{c|}{\makecell{\includegraphics[width = 0.1\textwidth]{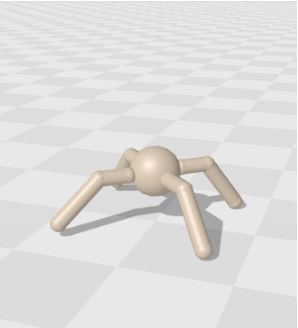}}}
    
    & \makecell{\includegraphics[width = 0.1\textwidth]{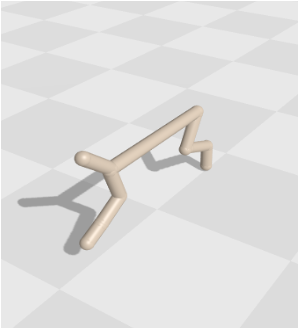}}
    
    & \multicolumn{2}{c|}{\makecell{\includegraphics[width = 0.1\textwidth]{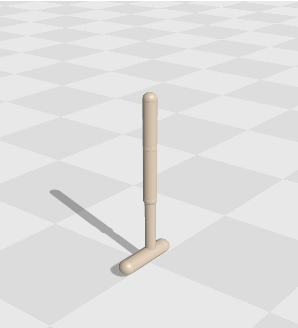}}}
    
    & \makecell{\includegraphics[width = 0.1\textwidth]{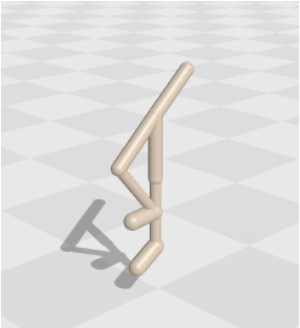}} \\

    \addlinespace[0.05cm]
    \midrule
    \addlinespace

    \makecell{\textsc{name}}
    & \makecell{\anttwo}
    & \makecell{\antthree}
    & \makecell{\halfcheetahtwo}
    & \makecell{\hoppertwo}
    & \makecell{\hopperthree}
    & \makecell{\walkertwo}
    \\
    
    \addlinespace[0.05cm]
    \midrule
    \addlinespace
     \makecell{\textsc{feature}}
     
     & \multicolumn{6}{c}{\makecell{Feet Contact Proportion}}\\
    
    \addlinespace[0.05cm]
    \midrule
    \addlinespace

    \makecell{\textsc{feature dim}}

    & \multicolumn{2}{c|}{4}
    
    & \makecell{2}
    
    & \multicolumn{2}{c|}{\makecell{1}}
    
    & \makecell{2} \\

    \addlinespace[0.05cm]
    \midrule
    \addlinespace
    
    \makecell{\textsc{Number of Parameters}}

    & \multicolumn{2}{c|}{10,312}
    
    & \makecell{5,766}
    
    & \multicolumn{2}{c|}{\makecell{5,123}}
    
    & \makecell{5,702} \\

    

    \addlinespace[0.05cm]
    \midrule
    \addlinespace[0.05cm]
    
    \makecell{\textsc{rewards}}&

    \begin{itemize}[leftmargin=*]
        \item Forward \newline velocity
        \item Energy consumption
    \end{itemize}
    
    &
    \begin{itemize}[leftmargin=*]
        \item $x$ velocity
        \item $y$ velocity
        \item Energy consumption
    \end{itemize}

    &
    \begin{itemize}[leftmargin=*]
        \item Forward \newline velocity
        \item Energy consumption
    \end{itemize}
    
    &
    \begin{itemize}[leftmargin=*]
        \item Forward \newline velocity
        \item Energy consumption
    \end{itemize}

    &
    \begin{itemize}[leftmargin=*]
        \item Forward \newline velocity
        \item Jumping height
        \item Energy consumption
    \end{itemize}

    &
    \begin{itemize}[leftmargin=*]
        \item Forward \newline velocity
        \item Energy consumption
    \end{itemize}
        
    
  \end{tabular}}
  \label{fig:experiments}
\end{table*}

\subsection{Evaluation Tasks}\label{section:tasks}

We evaluate our approach on six continuous control robotics tasks, which are summarised in \Cref{tab:tasks}. 
In these tasks, solutions correspond to the parameters of closed-loop neural network controllers 
which determine the torque commands given to each of the robot's joints.
In all tasks, the neural network controllers consist of two hidden layers of 64 neurons.  
The exact number of parameters depends on the observation and action dimensionalities of each environment, as summarised in \Cref{tab:tasks}.  
We use four robot morphologies from the Brax suite \cite{brax}.
In all of the tasks, the feature is the proportion of time that each of the robot's legs spends in contact with the ground.
Given a robot with $L$ legs (summarised in \Cref{tab:tasks}), the $L$-dimensional feature space is calculated as:

\begin{equation}
    \phi_i(x) = \frac{1}{T} \sum_{t=1}^T c_{i,t} \,\,\, \text{for}\,\,i=1, ..., L\,.
\end{equation}

Here, $c_{i,t}$ is $1$ when the leg is in contact with the ground at time step $t$, otherwise it is $0$. 
Using this characterisation, solutions that have diverse features will exhibit different gaits \cite{nature, mome}.

In four of the tasks (\anttwo, \halfcheetahtwo, \hoppertwo, \walkertwo) the aim is to maximise the forward velocity of the robot while minimising its energy consumption \cite{mome, mome-pgx}.
However, we also introduce two tri-objective \moqd environments: \antthree and \hopperthree.  
In the \antthree task, the objectives are the robot's $x$-velocity, $y$-velocity and energy consumption. 
Hence the goal is to discover controllers that lead to different gaits, and for each of these gaits to find controllers that travel in different directions while minimising the energy cost.
In the \hopperthree task, the rewards correspond to the robot's forward velocity, torso height and energy consumption. 
The corresponding \moqd goal is to therefore find solutions which have different gaits, and for each of these gaits to find controllers that make the hopper jump to different heights or travel forward, while minimising the energy cost.
These tasks were inspired by those in the \pgmorl paper \cite{prediction_guided_morl} but extended to be \moqd problems. We believe that they present interesting and realistic objectives, and also provide opportunity to compare \moqd algorithms on tasks with $m>2$.
\subsection{Baselines}
We compare \pcc to seven baselines: \pgx, \mome, \pga, \nsga and \spea.
Despite our best efforts, we were unable to implement \pgmorl as a \morl baseline, as the available code had compatibility issues with the QDax framework \cite{qdaxrepo} that we use for parallel evaluations of policies via Brax \cite{brax}, making \pgmorl more than 250 times slower than the other baselines.

\pgx and \mome are both \moqd algorithms so are straightforward to evaluate. 
However, \nsga and \spea are population-based multi-objective Evolutionary Algorithms (see \Cref{section:moeas}) which do not explicitly seek diversity and \pga is a mono-objective \qd algorithm (see \Cref{section:qd}).
To evaluate \pga we convert the multiple objectives into a single one by adding them.
To ensure that \pga had the same archive size as \moqd methods, if we use a grid of $k$ cells with maximum non-dominated front length of $|\mathcal{P}|$ for \moqd methods, we use $k \times |\mathcal{P}|$ cells for \pga.
To ensure that our comparisons are not confounded by differences in archive size, we also include an additional \pga baseline configured with the same number of grid cells as the MOQD methods (see \Cref{app:pga}).
By contrast, using a large population size for \nsga and \spea can reduce selection pressure and limit the number of generations, potentially hindering performance.
To address this, we run two versions of both baselines: (1) the \nsga and \spea baselines, which use population sizes as close to $k \times |\mathcal{P}|$ that were feasible within our GPU compute limits, and (2) the \nsgasmall and \speasmall baselines, which use the same batch size as the \qd baselines but run for more generations to match the total number of evaluations.

To report metrics for \pga, \nsga, \nsgasmall, \spea and \speasmall we use a passive archive with the same structure as the \moqd methods.
At each iteration, we fill the passive archive with solutions found by the algorithm and then calculate metrics on these archives.
Importantly, the passive archives do not interact within the primary algorithmic loop, ensuring that there is no effect on the behaviour of the baseline algorithms.

\subsection{Metrics}\label{section:metrics}
\label{sec:metrics}
We evaluate our method based on six metrics:
\begin{enumerate}
    \item The \textbf{\moqdscore} \cite{mome, mome-pgx} is the sum of the hypervolumes of the non-dominated fronts stored in the archive $\mathcal{A}$:
    \begin{equation*}
    \sum_{i=1}^{k} \Xi(\mathcal{P}_i), \,\,\text{where} \,\,\forall i, \mathcal{P}_i = \mathcal{P}(x \in \mathcal{A}|\Phi(x)\in C_i)
    \end{equation*}
    This metric aims to assess if an algorithm can find high-performing fronts, for a range of features.

    \item We introduce the \textbf{\moqdsparsityscore}, which we define as the average sparsity of each non-dominated front $\Xi(\mathcal{P}_i)$ of the archive:
    \begin{equation*}
    \frac{1}{k}\sum_{i=1}^{k} S(\mathcal{P}_i), \,\,\text{where} \,\,\forall i, \mathcal{P}_i = \mathcal{P}(x \in \mathcal{A}|\Phi(x)\in C_i)
    \end{equation*}
    We introduce this metric in \moqd settings as an attempt to measure whether, for each feature, the algorithm succeeds in finding a smooth trade-off of objective functions.

    \item The \textbf{\globalhypscore} is the hypervolume of the non-dominated front formed over all of the solutions in the archive (which we term the \textit{global non-dominated front}).
    The metric assesses the elitist performance of an algorithm.
    That is, the performance of solutions on the objective functions that are possible when disregarding the solution's feature.

    \item By the same reasoning as the \textbf{\moqdsparsityscore}, we also introduce the \textbf{\globalsparsity}, which is the sparsity of the non-dominated front formed over all of the solutions in the archive.

    \item We calculated the \textbf{\maxsumscores} of objective functions to compare our approach with traditional \qd algorithms which directly aim to maximise this.

    \item The \textbf{\coverage} is the proportion of cells in of an archive that are occupied. It reflects how many different features the algorithm is able to uncover (regardless of the performance of the solutions). Since all of the algorithms achieved a similar performance on this metric, we report the results for baselines and ablations in \Cref{app:coverage}
.
\end{enumerate}

The \moqd metrics (1 and 2) form evaluation methods that most closely align with assessing whether an algorithm achieves the \moqd goal given by \cref{eqn:moqd}.
The global metrics (3 and 4) assess the algorithms multi-objective performance, and allow for direct comparison with \mo baselines.
Since the sparsity metrics can be impacted by imbalanced scales, we run all of the baselines and find the minimum and maximum of the objectives seen across all of the baselines.
We then normalise all of the fitnesses based on these values, and report the sparsity metrics based of the final archives from the normalised fitness values.

\subsection{Hyperparameters}

All experiments were run for the same total budget of of $1,024,000$ evaluations.
In \pcc, \pgx, \mome, \nsga and \spea this corresponded to $4000$ iterations with a batch size of 256 evaluations per generation.
In \nsgasmall and \speasmall, we used a batch size of $1600$ which was the limit of our GPU and ran the algorithm for $640$ iterations.
We used CVT tessellation \cite{cvt} to create archives with $128$ cells, each with a maximum of non-dominated Front length of $50$.
For all experiments, we use a Iso+LineDD operator \cite{isoline} as the \ga variation operator, with $\sigma_1 = 0.005$ and $\sigma_2 = 0.05$.
This operator mutates a solution by adding isotropic Gaussian noise scaled by $\sigma_1$ and a directional perturbation along the vector between two individuals, scaled by $\sigma_2$, promoting both exploration and guided variation.
The reference points for each environment and the actor-critic training parameters were kept the same across all algorithms and are provided \Cref{app:refpoints} and \Cref{app:pghyperparams}.

\begin{figure*}[ht!]
\centering
\includegraphics[width=\textwidth]{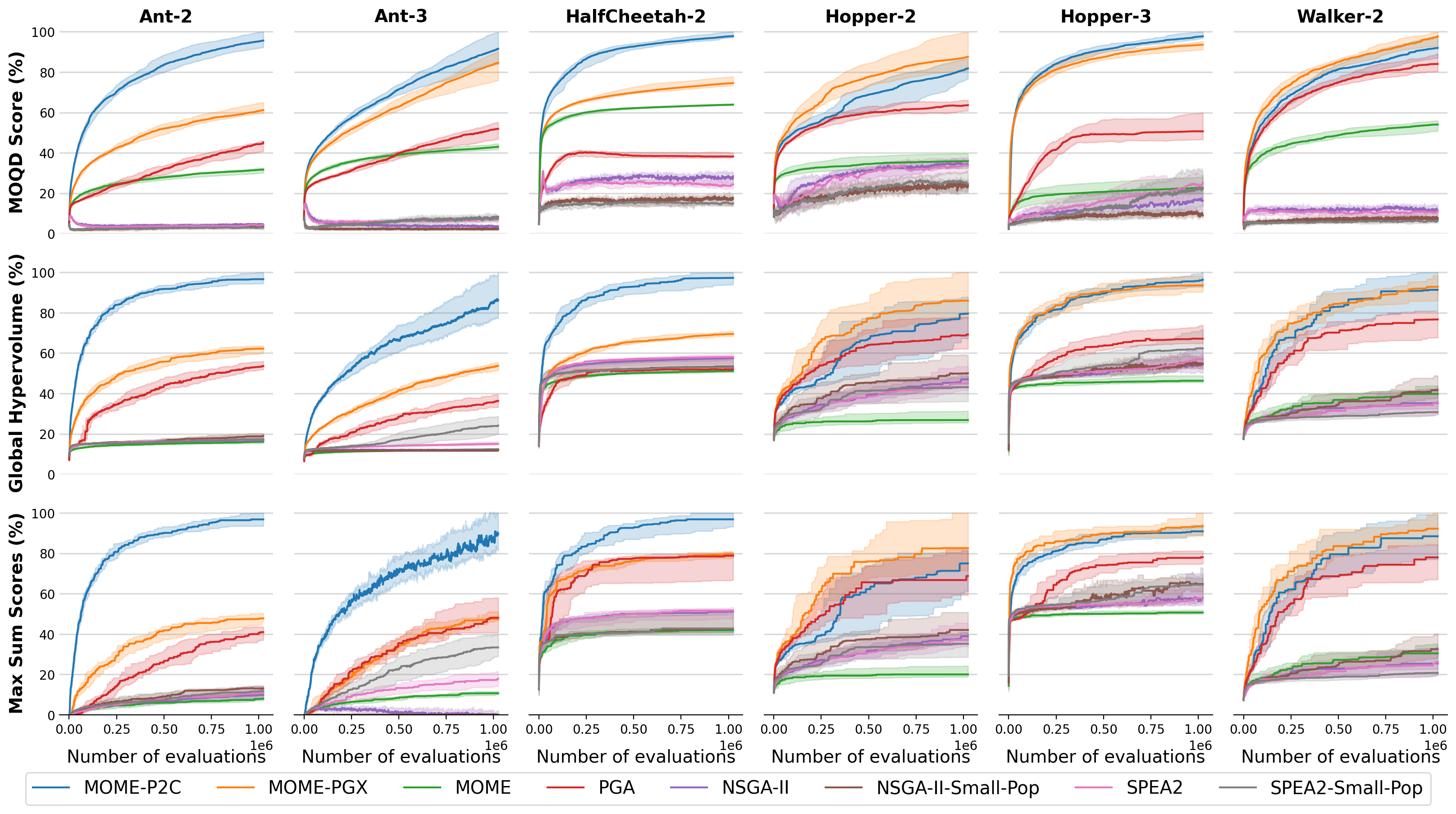}
\caption{\moqdscore, \globalhypscore and \maxsumscores (\cref{sec:metrics}) for \pcc compared to all baselines across all tasks. Each experiment is replicated 20 times with random seeds. The solid line is the median and the shaded area represents the first and third quartiles.}
\label{fig:main_results}
\end{figure*}

\section{Results}
In this section, we present the results for all baselines. Each experiment is replicated \replications times with random seeds.
We report $p$-values based on the Wilcoxon–Mann–Whitney $U$ test with Holm-Bonferroni correction to ensure statistical validation of the results \cite{wilcoxon1992individual, holm_bonf}.

\begin{figure}[ht!]
\centering
./\includegraphics[width=0.8\linewidth]{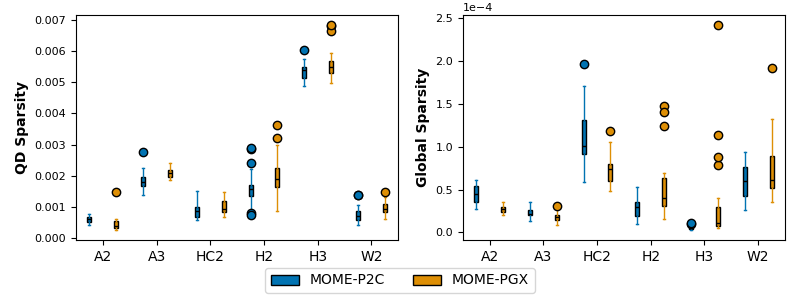}
\caption{Boxplots to display sparsity metrics calculated on the final archive of \pcc and \pgx over \replications replications. The labels A2, A3, HC2, H2, H3 and W2 correspond to the \anttwo, \antthree, \halfcheetahtwo, \hoppertwo, \hopperthree and \walkertwo environments respectively.}
\label{fig:sparsity_box}
\end{figure}

\begin{figure*}[ht!]
\centering
\includegraphics[width=\textwidth]{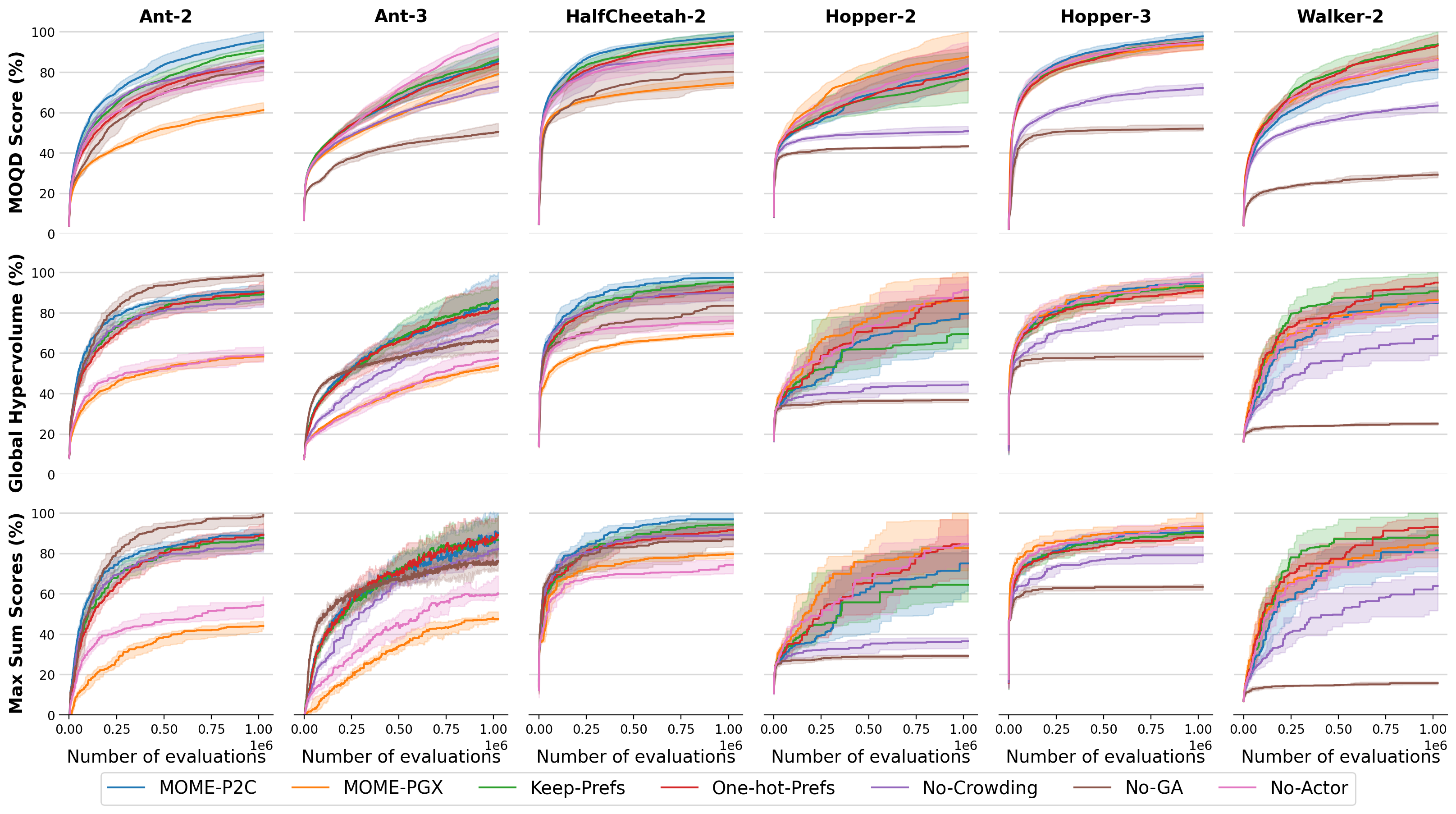}
\caption{\moqdscore (\cref{sec:metrics}) for \pcc compared to all ablations across all tasks. Each experiment is replicated \replications times with random seeds. The solid line is the median and the shaded area represents the first and third quartiles.}
\label{fig:ablation_results}
\end{figure*}

\subsection{Main Results}
\label{section:results}
The experimental results presented in \cref{fig:main_results} demonstrate that \pcc outperforms or matches all baselines on all tasks and all metrics.
\pcc achieves a significantly higher \moqdscore than all baselines on \anttwo, \antthree, \halfcheetahtwo and \hopperthree ($p < 0.02$).
\pcc matches the \moqdscore of \pgx, the previous state-of-the-art, on the remaining environments \hoppertwo and \walkertwo.
We include archive visualisations for environments with two-dimensional features in \Cref{app:repertoires} to further visualise these results.
Crucially, in the tasks where \pcc does not markedly outperform \pgx on \moqdscore, it still attains lower \moqdsparsityscore (\cref{fig:sparsity_box}), indicating that it achieves smoother array of trade-offs for each feature.

\pcc also outperforms or matches all baselines on the \globalhypscore metric.
We include visualisations of the global non-dominated Fronts obtained by representative runs of each of the algorithms in \Cref{app:pfs}.
\pcc outperforms \nsga, \nsgasmall, \spea  and \speasmall on the \globalhypscore in all environments, algorithms that specifically aim to maximise this metric.
We hypothesise that this result is due to the policy-gradient variations of \pcc, which improve the search capabilities of these algorithms in high-dimensional search spaces with neural network controllers. 
However, \pcc still outperforms or matches the \globalhypscore of the other baselines which use policy-gradient variations (\pga and \pgx), demonstrating the benefits of using preference-conditioned updates.
Importantly, in the environments where \pcc matches \pgx on \globalhypscore (\hoppertwo, \hopperthree, \walkertwo), it still achieves a lower \globalsparsity as illustrated in \Cref{fig:sparsity_box}.

Furthermore, \pcc achieves a better \maxsumscores than \pga across all tasks ($p < 10^{-5}$) except \hoppertwo and \walkertwo where it still shows improved but not statistically significant performance.
This highlights that even if a specific reward preference is available, leveraging multiple objectives may still result in better exploration and higher returns compared to single-objective baselines.
Furthermore, we see that \pcc achieves better performance on tri-dimensional tasks,  affirming not only its computational efficiency but also its scalability in handling more complex tasks with an increased number of objectives.

\subsection{Ablations}
\label{section:ablation_results}

\subsubsection{Ablation studies}
In our ablation studies, we evaluate \pcc against \pgx together with five distinct modifications to understand the contribution of each component in \pcc. These ablations include:
\begin{itemize}
    \item \textbf{\keeppref}: \pcc with a policy-gradient variation operator that keeps the preference of the parent instead of sampling a new preference.
    \item \textbf{\onehot}: \pcc with a policy-gradient variation operator that uses equal batch sizes of one-hot preferences. The \onehot ablation is the same as \pgx except with a preference-conditioned actor-critic network, rather than separate networks.
    \item \textbf{\noactor}: \pcc without the actor injection mechanism. Instead of generating 64 policy-gradient offspring and 64 actor-injection offspring at each generation, \noactor produces 128 policy-gradient offspring.
    \item \textbf{\noga}: \pcc without any genetic mutations. At each generation, there are 128 policy-gradient offspring and 128 actor-injection offspring. 
    \item \textbf{\nocrowding}: \pcc without crowding mechanisms, that employs uniform selection and replacement.
\end{itemize}

\subsubsection{Results}\label{sec:ablation_results}

The results from our ablation studies provide a deeper understanding of the individual components contributing to \pcc's effectiveness, we include further archive visualisations in \Cref{app:repertoires}.
Firstly, we examine the modifications in preference sampling strategies, as explored in the \keeppref and \onehot ablations.
\pcc matches or outperforms the \keeppref ablation, which retains the parent's preference in the policy-gradient variation operator, across all tasks except \walkertwo. 
Similarly, the \onehot ablation, employing one-hot preferences in equal batch sizes has no statistical difference or is worse compared to \pcc across all tasks in all metrics except \walkertwo.  
These results highlight that \pcc is robust to the preference-sampling strategy in most environments.
However, the results on \walkertwo suggests that, in some environments, alternative sampling strategies may be optimal.
We leave this as a promising avenue for future research.
Additionally, we note that \onehot outperforms or matches \pgx across all metrics, in all environments, highlighting that simply using the preference-conditioned critic leads to improved performance.

The \nocrowding ablation, where \pcc operates without its crowding mechanisms, significantly underperforms compared to \pcc across all tasks ($p < 10^{-4}$).
This highlights that the crowding-based selection and addition mechanism is crucial for guiding the algorithm towards more diverse and high-quality solutions.
Likewise, \pcc also outperforms the \noga ablation in every environment except for \anttwo, where the \noga ablation environment achieves a higher \globalhypscore and \maxsumscores.
We hypothesise that since the Ant robot has higher dimensionality than other morphologies, in this environment the high-dimensionality of the search space requires more gradient-based search power.
However, the \noga has a worse performance across all other metrics in all of the other environments, particularly \hoppertwo, \hopperthree and \walkertwo, highlighting that in general the genetic variations are an essential component of the \pcc algorithm.

Finally, we consider the \noactor ablation, which removes the actor injection mechanism from \pcc.
In the \anttwo and \halfcheetahtwo ($p < 10^{-4}$), \pcc markedly outperforms the \noactor ablation, suggesting that the actor injection mechanism is essential in these tasks.
Moreover, \noactor either matches or falls behind the full \pcc model in all other tasks and all other metrics, except for the \moqdscore in \antthree.

\section{Conclusion}
In this paper, we have introduced a novel algorithm, \textbf{\pcc}, which represents a significant advancement in the field of Multi-Objective Quality-Diversity (MOQD) optimisation. Our experiments and ablation studies have demonstrated \pcc's ability to balance multiple objectives effectively, outperforming existing state-of-the-art methods in challenging continuous control environments. 

One of the key strengths of \pcc is its use of preference-conditioned policy gradient mutations, which not only enhance the exploration process but also ensures an even distribution of solutions across the non-dominated front. This approach addresses the limitations of \pgx that struggled with scalability and efficiency. Furthermore, \pcc's ability to perform well in tri-dimensional tasks highlights its scalability and adaptability to more complex and realistic scenarios.

Beyond these results, a limitation of our study is that experiments were restricted to standard Brax locomotion benchmarks with up to three objectives; extending evaluation to higher-dimensional and non-robotics tasks remains an important avenue for future work.
Additionally, a key limitation of \pcc lies in how preferences are currently sampled.
While the algorithm is generally robust across environments, certain tasks, such as \walkertwo, demonstrate that naive or uniform preference sampling may be suboptimal.
This suggests that dynamic or adaptive sampling strategies could further improve performance, especially in environments with complex or shifting trade-offs.
The exploration of using models to predict which preference will lead to the largest hypervolume gain \cite{prediction_guided_morl} presents an exciting direction for further research.

Finally, while our parameter comparisons in Appendix~\ref{app:pghyperparams} demonstrate that \pcc is more memory-efficient than \pgx, scaling to higher-dimensional objective spaces introduces additional challenges. 
In particular, we expect random preference sampling to become increasingly ineffective as dimensionality grows and, as noted in Section 4.1, crowding distance is known to lose effectiveness in many-objective optimisation \cite{zheng2024truthful, koppen2007substitute}. 
Addressing these issues through adaptive preference sampling, scalable diversity-preservation mechanisms, and architectural innovations represents a key line of future work in extending the generality of \pcc.  


\section*{Acknowledgements}
This work was supported by PhD scholarship funding for Hannah from InstaDeep.

\bibliographystyle{ACM-Reference-Format}
\bibliography{bibliography}

\clearpage
\appendix

\section{Coverage Results}\label{app:coverage}

\begin{figure*}[ht!]
    \centering
    \includegraphics[width=\textwidth]{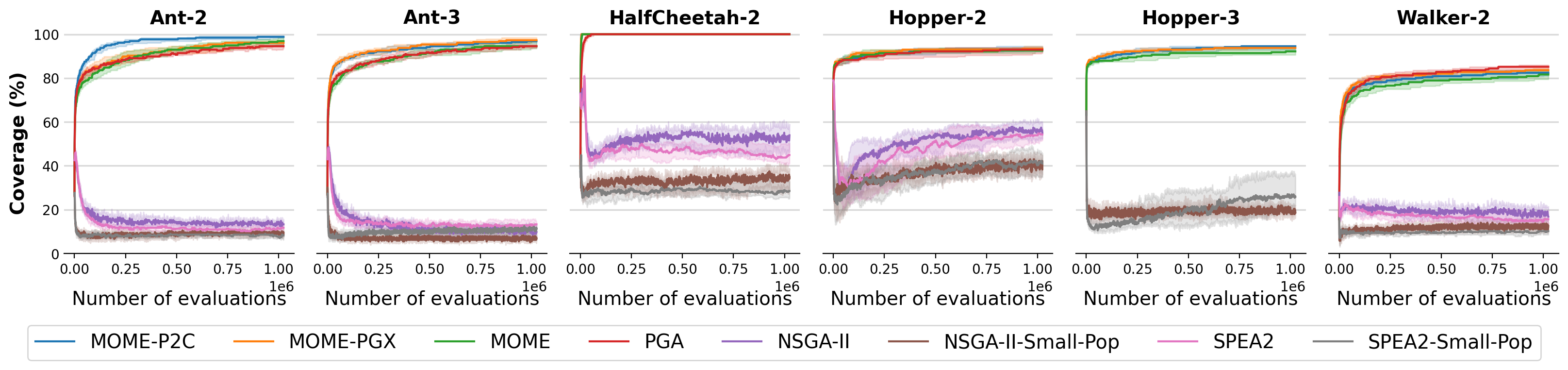}
    \caption{Median coverage performance of \replications seeds, the shaded regions show the inter-quartile range.}
    \label{fig:coverage}
\end{figure*}

\begin{figure*}[ht!]
    \centering \includegraphics[width=\textwidth]{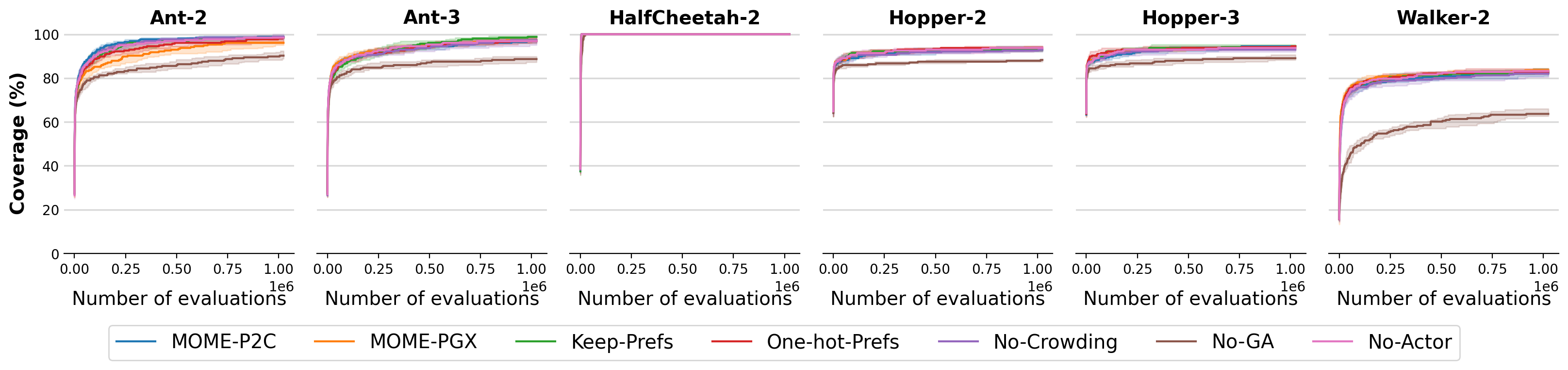}
    \caption{Median coverage performance of \replications seeds, the shaded regions show the inter-quartile range.}
    \label{fig:abations_coverage}
\end{figure*}

\Cref{fig:coverage} and \Cref{fig:abations_coverage} present the coverage results for all of the baseline and ablations algorithms respectively. As expected, all Quality-Diversity algorithms achieve a higher coverage score than the MOEA baselines, as they explicitly seek diverse solutions.

\section{Hypervolume Reference Points} \label{app:refpoints}
Table \ref{tab:refpoints} presents the reference points used to calculate the hypervolume metrics in each of the tasks. 
The same reference points were used for all of the experiments.

\begin{table}[ht!]
    \centering
        \caption{Reference points}
    \begin{tabular}{l | l }
        \toprule
    Ant-2 & [-350, -4500] \\
    Ant-3 & [-1200, -1200, -4500] \\
    HalfCheetah-2 & [-2000, -800] \\
    Hopper-2 & [-50, -2] \\
    Hopper-3 & [-750, -3, 0]\\
    Walker-2 & [-210, -15] \\
        \bottomrule
    \end{tabular}      	
    \label{tab:refpoints}
\end{table}

\newpage
\section{Policy Gradient Hyperparameters}\label{app:pghyperparams}

Table \ref{tab:pghyps} presents all of the policy gradient hyperparameters that are used for our algorithms.
All hyperparameters were kept the same for each task and for all algorithms which used PG variations.

\begin{table}[ht!]
    \centering
        \caption{Policy Gradient Network Hyperparameters}
    \begin{tabular}{l | l }
        \toprule
    Replay buffer size & 1,000,000 \\
    Critic training batch size & 256\\
    Critic layer hidden sizes & $[256, 256]$\\
    Critic learning rate &  $3\times 10^{-4}$\\
    Actor learning rate &  $3\times 10^{-4}$\\
    Policy learning rate &  $1\times 10^{-3}$\\
    Number of critic training steps & 300\\
    Number of policy gradient training steps & 100\\
    Policy noise & 0.2\\
    Noise clip & 0.2\\
    Discount factor & 0.99\\
    Soft $\tau$-update proportion &  0.005\\
    Policy delay & 2\\
        \bottomrule
    \end{tabular}      	
    \label{tab:pghyps}
\end{table}

In addition to hyperparameters, we report the number of parameters required for actor and critic networks in \pcc compared to \pgx across all environments. 
These values directly support the argument in the main text regarding the improved memory efficiency of \pcc. 

\begin{table}[h]
\centering
\caption{Number of parameters in total for Critic and Actor Networks for \pcc and \pgx in each environment.}
\begin{minipage}{0.48\linewidth}
\centering
\begin{tabular}{lcc}
\toprule
 & \pcc & \pgx \\
\midrule
\anttwo        & 182,788 & 362,500\\
\antthree      & 183,814 & 543,750\\
\halfcheetahtwo & 146,436 & 289,796\\
\hoppertwo     & 141,316 & 279,556\\
\hopperthree   & 142,342 & 419,334\\
\walkertwo     & 145,924 & 288,772\\
\bottomrule
\end{tabular}
\subcaption{Critic Networks}
\end{minipage}
\hfill
\begin{minipage}{0.48\linewidth}
\centering
\begin{tabular}{lcc}
\toprule
 & \pcc & \pgx \\
\midrule
\anttwo         & 90,376  & 180,752 \\
\antthree       & 90,376  & 271,128 \\
\halfcheetahtwo & 72,198  & 144,396 \\
\hoppertwo      & 69,635  & 139,270 \\
\hopperthree    & 69,635  & 208,905 \\
\walkertwo      & 71,942  & 143,884 \\
\bottomrule
\end{tabular}
\subcaption{Actor Networks}
\end{minipage}
\end{table}

As shown, \pcc consistently requires roughly half the number of parameters compared to \pgx across all environments, confirming its memory efficiency advantage. 
This reduction holds for both actor and critic networks, and becomes more pronounced as the number of objectives increases (e.g., \antthree and \hopperthree).  
The difference in parameter counts between \anttwo and \antthree, and similarly between \hoppertwo and \hopperthree, illustrates how scaling to additional objectives affects the two approaches. 
For \pgx, the number of parameters grows roughly linearly with the number of objectives, since a separate actor–critic pair is maintained for each objective. 
In contrast, \pcc shows only a marginal increase in parameters between the two- and three-objective tasks, as the same networks are reused and only the input dimensionality grows. 
This highlights that while both methods scale with the number of objectives, \pcc’s architecture is more memory-efficient and thus better suited to higher-dimensional problems.  

\begin{figure}[h]
    \centering
    \includegraphics[width=\textwidth]{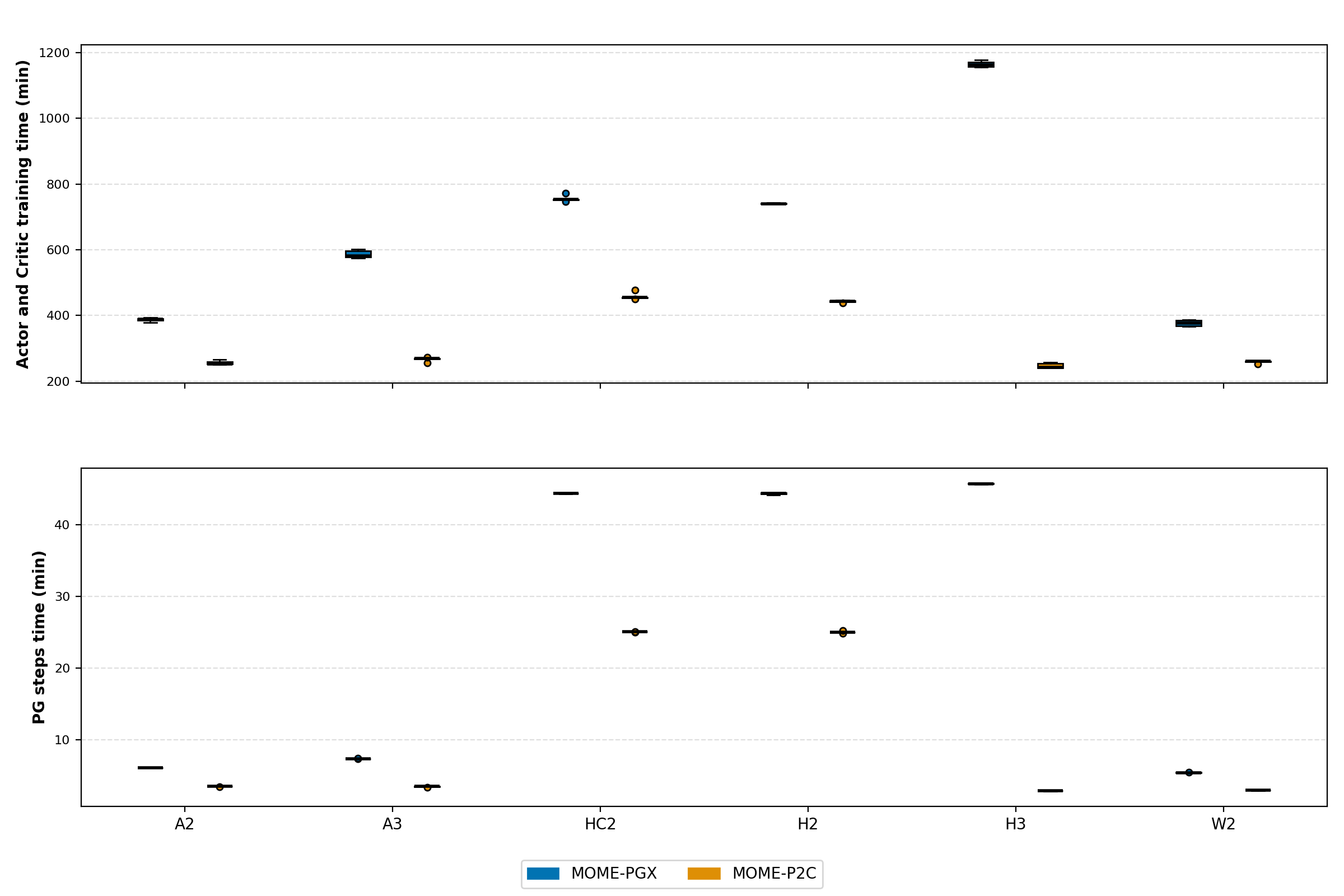}
    \caption{Wall-clock training time (in minutes) for policy-gradient mutation steps and actor–critic training in each environment. 
    \pcc consistently requires less time than \pgx across all tasks. 
    Note: timings were collected from un-jitted runs for accurate measurement; actual JIT-compiled training is substantially faster.}
    \label{fig:wallclock}
\end{figure}

To further support our claims of computational efficiency, \cref{fig:wallclock} reports the wall-clock time of policy-gradient mutation steps and actor–critic training for \pcc and \pgx across all environments. 
We find that \pcc consistently requires less training time than \pgx in every environment, demonstrating its improved computational efficiency. 
To obtain accurate timing measurements, we disabled JIT compilation in the QDax framework, as JIT fusion obscures intermediate operation timings. 
As a result, the reported values are significantly longer than those observed in practice, but they provide a fair relative comparison across methods. 
In typical JIT-compiled runs, each environment for \pcc completes in under one hour depending on the task. 
Because un-jitted runs are substantially slower, we report results for five random seeds per environment. 
Overall, the lower wall-clock times and reduced parameter counts together confirm that \pcc achieves both computational and memory efficiency over \pgx.

\section{Ablation – Effect of Archive Size for PGA}\label{app:pga}

\begin{figure}[h]
    \centering
    \includegraphics[width=0.95\textwidth]{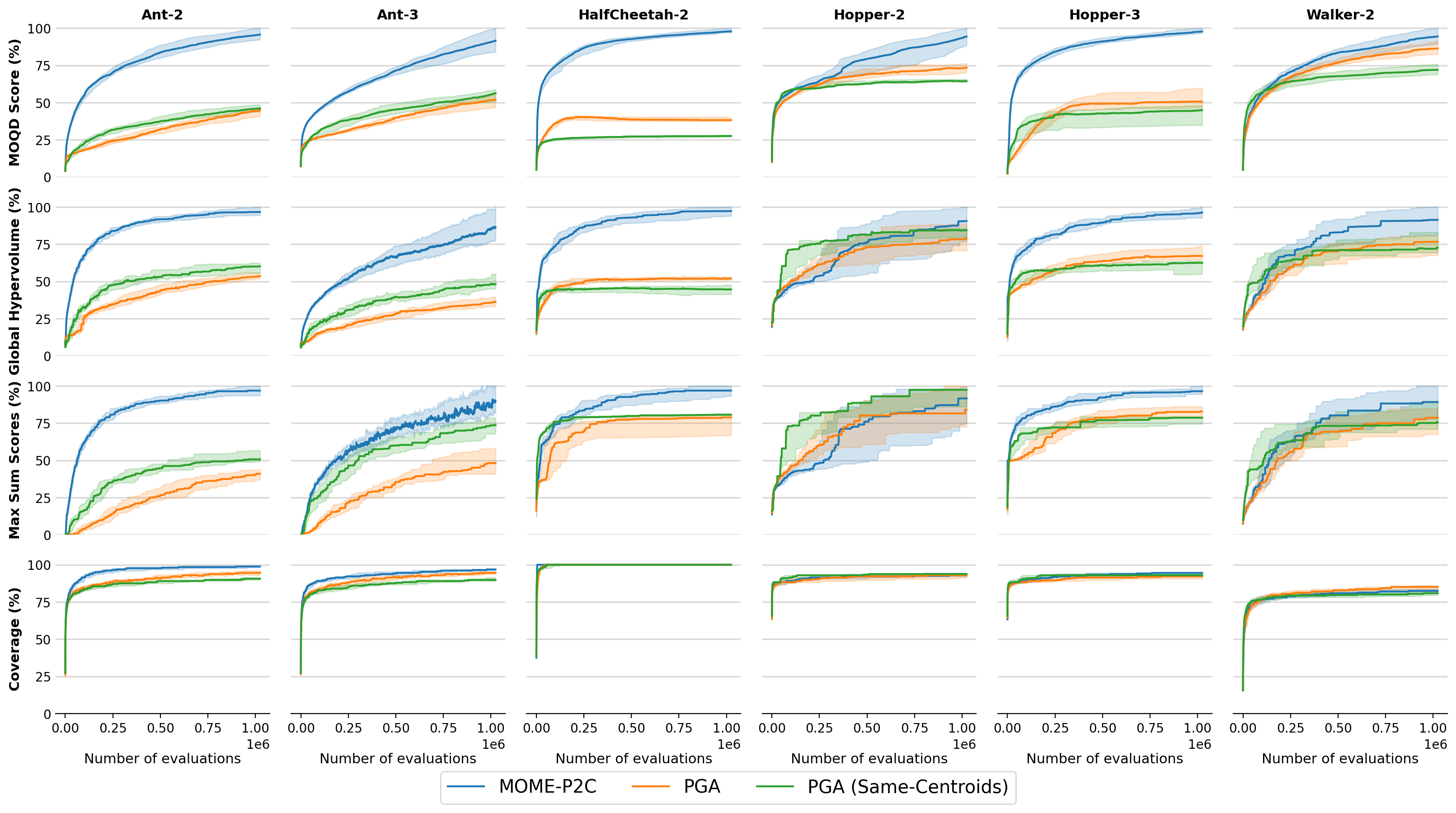}
    \caption{Comparison between \pcc and \pga when both use the same number of grid cells $k$.  
    Each curve shows the mean and interquartile range over 20 runs.  
    Results demonstrate that \pcc consistently outperforms or matches \pga across environments, even when controlling for archive size.}
    \label{fig:pga_samecells}
\end{figure}

In the main text, we compared \pcc to \pga using an archive size of $k \times |\mathcal{P}|$ cells, to to ensure a fair comparison in terms of archive capacity.
However, having a large number of cells may decrease the selection pressure of the algorithm and therefore negatively impact the performance of \pga. 
We therefore reran \pga with the same number of grid cells $k$ as the MOQD methods.  
The results are reported in \cref{fig:pga_samecells}.

We observe that \pcc continues to achieve a higher \moqdscore across all tasks ($p < 0.001$).
\pcc also significantly outperforms \pga on the \maxsumscores metrics in most environments, except \walkertwo and \hoppertwo where results are not statistically significant.
The same trend is observed for the \globalhypscore, where \pcc significantly outperforms \pga on \anttwo, \antthree, and \halfcheetahtwo ($p < 10^{-4}$), as well as \hopperthree ($p < 0.01$).  
Taken together, these results confirm that the observed improvements are not merely an artefact of archive size: even when controlling for the number of cells, \pcc consistently outperforms or matches \pga across environments.

\section{Repertoire Visualisations}
\label{app:repertoires}

\begin{figure}[h]
    \centering
    \includegraphics[width=0.9\textwidth]{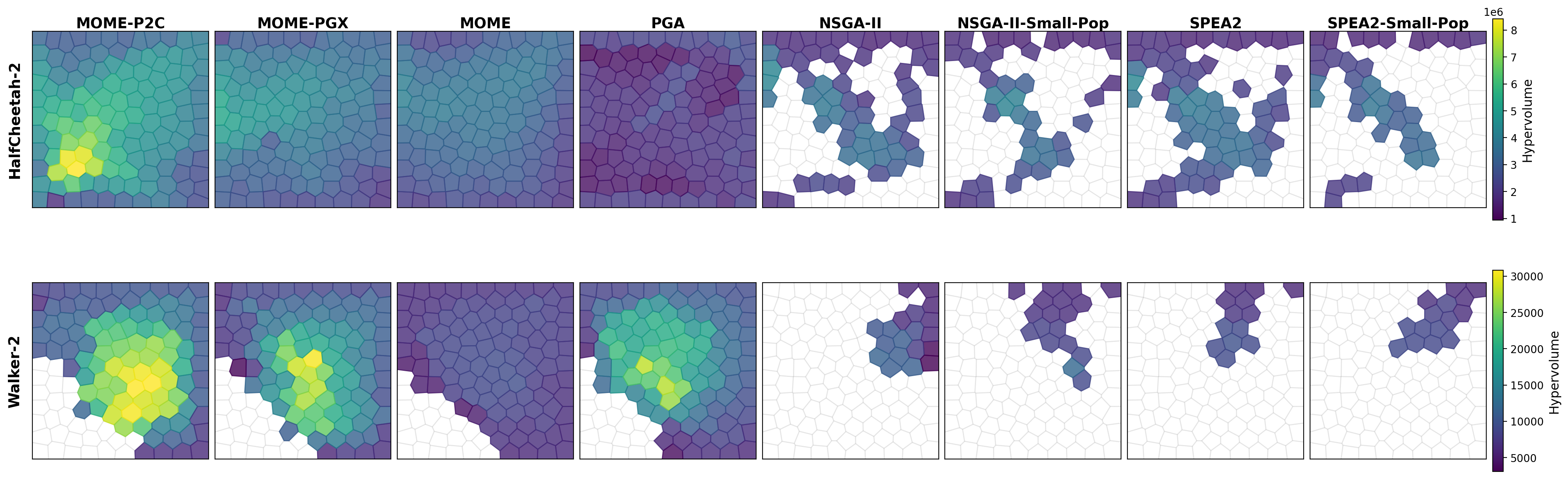}
    \caption{Representative repertoires for the main baselines in experiments with two-dimensional features. Colours indicate the hypervolume of the non-dominated set in each cell. \pcc achieves broader coverage and higher-quality solutions compared to other baselines.}
    \label{fig:repertoires_baselines}
\end{figure}

\begin{figure}[h]
    \centering
    \includegraphics[width=0.9\linewidth]{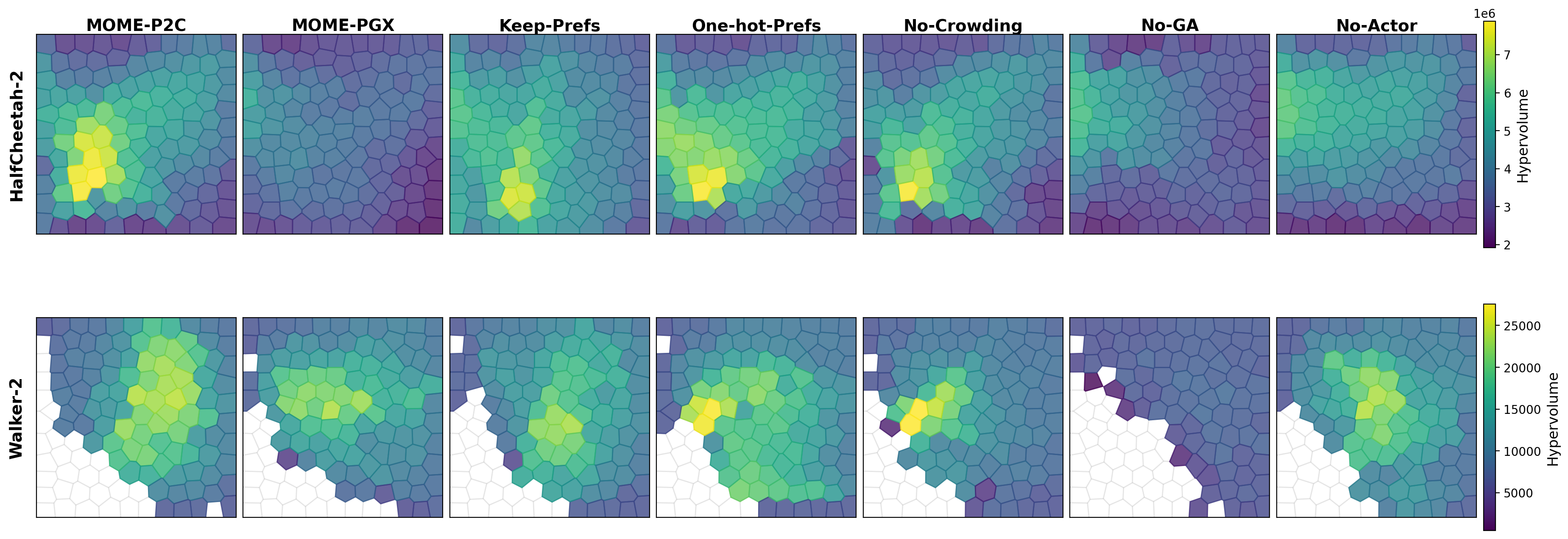}
    \caption{Representative repertoires for the ablation studies in experiments with two-dimensional features. Each ablation shows a noticeable drop in coverage and/or quality compared to \pcc, consistent with the quantitative results in the main paper.}
    \label{fig:repertoires_ablations}
\end{figure}

To provide additional insight into the behaviour of the algorithms, we visualise the final repertoires in the feature space for environments with two-dimensional feature descriptors.  
Figures~\ref{fig:repertoires_baselines} and~\ref{fig:repertoires_ablations} show representative seeds for each baseline and ablation, respectively.  

In Figure~\ref{fig:repertoires_baselines}, we observe that \pcc produces larger and denser archives with non-dominated fronts with higher hypervolume than the other baselines.  
Compared to population-based MOEAs such as \nsga and \spea, which often leave large regions of the feature space unexplored, \pcc and other \moqd algorithms cover the space more uniformly.  
We also note that that the other baselines achieve good coverage, but fail to discover as high-performing fronts as \pcc.  

Figure~\ref{fig:repertoires_ablations} aligns with the quantitative results in Section~\ref{section:results}, where we observe that in the \halfcheetahtwo tasks \pcc performs much better than the ablations, but the ablations are competitive in \walkertwo environment.

\section{Global Non-Dominated Fronts}\label{app:pfs}

\begin{figure*}[ht!]
    \centering
    \includegraphics[width=0.9\textwidth]{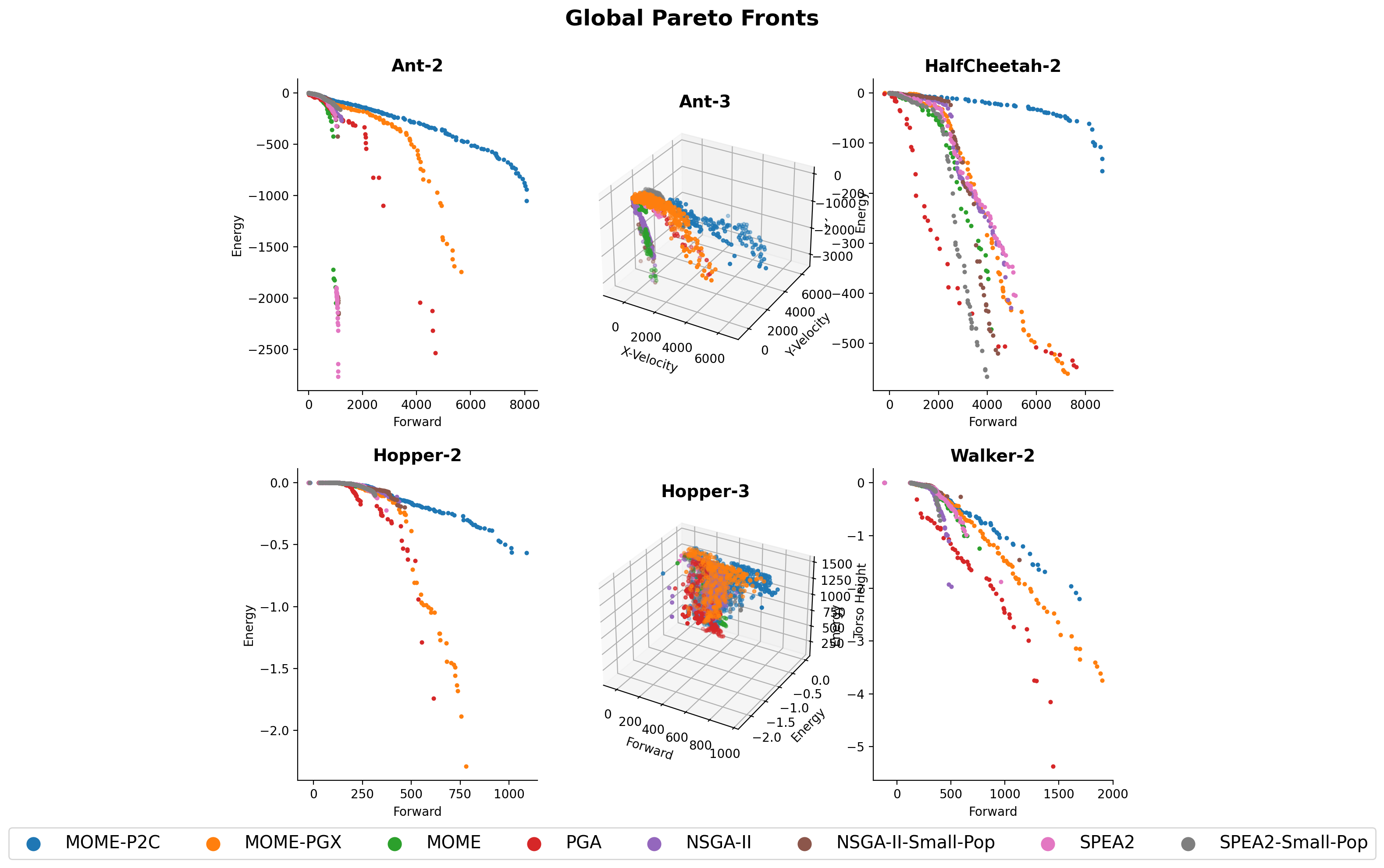}
    \caption{Visualisations of global non-dominated Fronts obtained by representative runs of each of the baseline algorithms.}
    \label{fig:global_pf}
\end{figure*}

\begin{figure*}[ht!]
    \centering
    \includegraphics[width=0.9\textwidth]{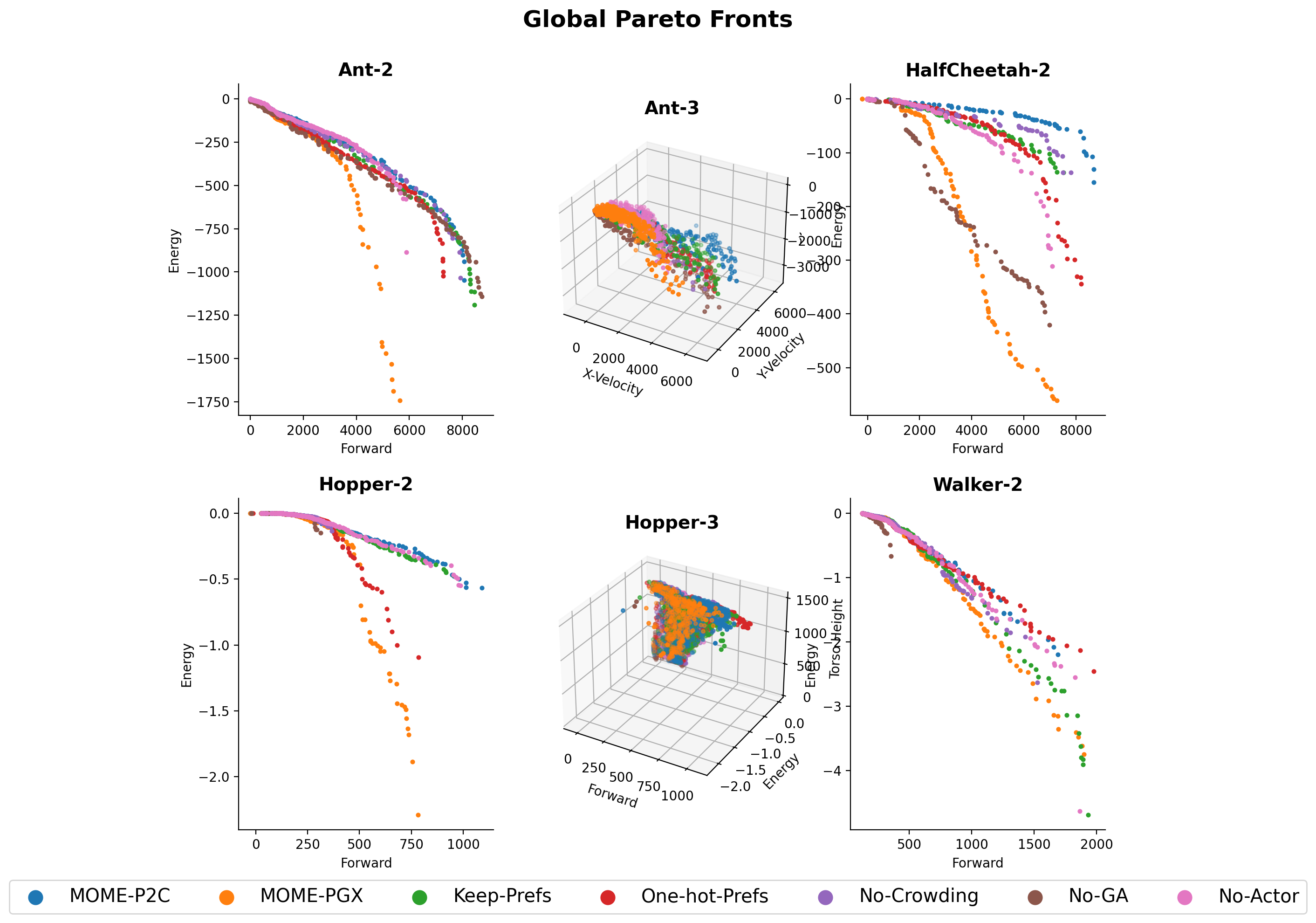}
    \caption{Visualisations of global non-dominated Fronts obtained by representative runs of each of the ablation algorithms.}
    \label{fig:ablations_global_pf}
\end{figure*}

\Cref{fig:global_pf} and \Cref{fig:ablations_global_pf} presents the visualisations of the global non-dominated fronts obtained by each of the baseline and ablation algorithms.

\end{document}